\documentclass[sigconf]{aamas} 
\usepackage{balance}
\newcommand\set[1]{\{ #1\}}
\newcommand{\mb}{\mathbf}
\usepackage{amsmath}
\usepackage{subcaption}
\usepackage{tikz}
\usetikzlibrary{math}
\usetikzlibrary{plotmarks}
\usepackage{booktabs}
\usepackage{hyperref}
\usepackage[capitalise]{cleveref}
\usepackage{xcolor}
\usepackage{amsfonts}
\usepackage{multirow}
\usepackage{paralist}

\usepackage{comment}

\usepackage{xspace}
\newcommand*{\eg}{e.g.,\@\xspace}
\newcommand*{\ie}{i.e.,\@\xspace}

\usepackage{xcolor}
\usepackage{xspace}
\newcounter{commentCounter}
\newif\iftrvar
\trvartrue
\iftrvar

\newcommand{\tim}[1]{{\small \color{red} \refstepcounter{commentCounter}\textsf{[TR]$_{\arabic{commentCounter}}$:{#1}}}}
\newcommand{\minqi}[1]{{\small \color{purple} \refstepcounter{commentCounter}\textsf{[MJ]$_{\arabic{commentCounter}}$:{#1}}}}
\newcommand{\zhengyao}[1]{{\small \color{blue} \refstepcounter{commentCounter}\textsf{[ZJ]$_{\arabic{commentCounter}}$:{#1}}}}
\newcommand{\pasquale}[1]{{\small \color{orange} \refstepcounter{commentCounter}\textsf{[PM]$_{\arabic{commentCounter}}$:{#1}}}}
\else

\newcommand{\tim}[1]{}
\newcommand{\minqi}[1]{}
\newcommand{\pasquale}[1]{}
\newcommand{\zhengyao}[1]{}
\fi

\begin{document}
\title[Grid-to-Graph: Flexible Spatial Relational Inductive Biases for Reinforcement Learning]{Grid-to-Graph: Flexible Spatial Relational Inductive Biases \\ for Reinforcement Learning}

\setcopyright{ifaamas}
\acmConference[AAMAS '21]{Proc.\@ of the 20th International Conference on Autonomous Agents and Multiagent Systems (AAMAS 2021)}{May 3--7, 2021}{London, UK}{U.~Endriss, A.~Now\'{e}, F.~Dignum, A.~Lomuscio (eds.)}
\copyrightyear{2021}
\acmYear{2021}
\acmDOI{}
\acmPrice{}
\acmISBN{}

\acmSubmissionID{229}

\author{Zhengyao Jiang}
\affiliation{
  \department{Centre for Artificial Intelligence}
  \institution{University College London}}
\email{zhengyao.jiang.19@ucl.ac.uk}

\author{Pasquale Minervini}
\affiliation{
  \department{Centre for Artificial Intelligence}
  \institution{University College London}}
\email{p.minervini@cs.ucl.ac.uk}

\author{Minqi Jiang}
\affiliation{
  \department{Centre for Artificial Intelligence}
  \institution{University College London}}
\email{msjiang@cs.ucl.ac.uk}

\author{Tim Rocktäschel}
\affiliation{
  \department{Centre for Artificial Intelligence}
  \institution{University College London}}
\email{t.rocktaschel@cs.ucl.ac.uk}

\begin{abstract} 
Although reinforcement learning has been successfully applied in many domains in recent years, we still lack agents that can systematically generalize.  While relational inductive biases that fit a task can improve generalization of RL agents, these biases are commonly hard-coded directly in the agent's neural architecture. In this work, we show that we can incorporate relational inductive biases, encoded in the form of relational graphs, into agents. Based on this insight, we propose Grid-to-Graph (GTG), a mapping from grid structures to relational graphs that carry useful spatial relational inductive biases when processed through a Relational Graph Convolution Network (R-GCN). We show that, with GTG, R-GCNs generalize better both in terms of in-distribution and out-of-distribution compared to baselines based on Convolutional Neural Networks and Neural Logic Machines on challenging procedurally generated environments and MinAtar. Furthermore, we show that GTG produces agents that can jointly reason over observations and environment dynamics encoded in knowledge bases.
\end{abstract}
\keywords{Relational Inductive Bias; Reinforcement Learning; Graph Neural Network}
\maketitle

\section{Introduction} \label{chapterlabel1}
Reinforcement Learning (RL) has seen many successful applications in recent years. 
However, developing agents that can systematically generalize to out-of-distribution observations remains an open challenge~\citep{Zhang2018,cobbe2020,Farebrother2018}.
Relational inductive biases are considered important in promoting both in-distribution and systematic generalization, in supervised learning and RL settings~\citep{Zhang2018,zambaldi2018deep,Wang2018}.

\citet{battaglia2018relational} define relational inductive biases as constraints on the relationships and interactions among entities in a learning process.
Traditionally, relational inductive biases have been hard-coded in an agent's neural network architecture.
By tailoring the connections between neurons and applying different parameter sharing schemes, architectures can embody various useful inductive biases.
For example, convolutional layers~\citep{krizhevsky2012imagenet} exhibit locality and spatial translation equivariance~\cite{PuschelM08a}, a particularly useful inductive bias for computer vision, as the features of an object should not depend on its coordinates in an input image.
Similarly, recurrent layers~\citep{hochreiter1997long} and Deep Sets respectively exhibit time translation %equivariance
and 
%
%Deep Sets, 
permutation equivariance~\citep{zaheer2017deep, battaglia2018relational}.
%
%\minqi{Not clear what role this passage serves} However, as novel architecture designs typically make use of multiple inductive biases coupled together, it is difficult for researchers to investigate relational inductive bias in a principled way or to incorporate new ones easily.
%
% allowing the learning process to search through a family of relational inductive biases, without modifying the underlying neural network architecture.
%

In this work, we introduce a unified graph-based framework %based on graphs 
that allows us to express %many common 
several useful relational inductive biases %in a common formalism.
using the same formalism.
Specifically, we frame the computation graph underlying a neural architecture as a directed multigraph with parameter sharing groups denoted by common edge labels connecting shared parameters in each group.
This formalization allows us to define specific inductive biases as comprising rules that generate edges and edge labels. 
The computation is then implemented by Relational Graph Convolutional Networks \citep[R-GCNs,][]{schlichtkrull2018modeling}, a type of Graph Neural Networks~\citep[GNNs,][]{zhou2018graph} that dynamically construct a computation graph based on a relational graph.

We make use of this formalism to introduce Grid-to-Graph (GTG), a mapping from grid structures of discrete 2D observations to relational graphs, based on a set of \emph{relation determination rules} that generate effective spatial relational inductive biases.
Given a feature map with entities (nodes) arranged in a lattice where each entity corresponds to a feature vector, the relations %embodied 
encoded by GTG constrain the flow of information between these entity feature vectors when these features are processed by R-GCNs. We refer to the resulting approach as R-GCN-GTG.

\begin{figure*}[t]
    \centering
    \includegraphics[width=\textwidth]{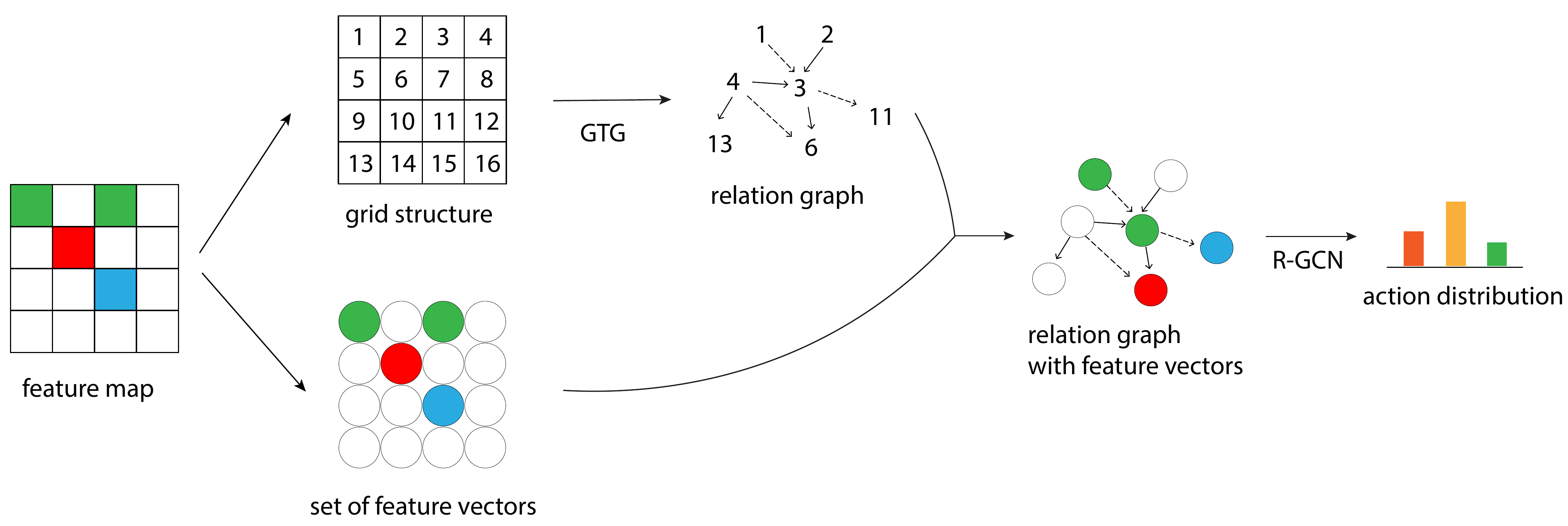}
    \caption{A high-level overview of GTG-R-GCN. We abstract away the grid structure of the feature map and turn observations into a spatial relational graph with GTG. The vectors of the feature map are attached to nodes in the relational graph. We then use an R-GCN to reason over the relational graph and node feature vectors to produce an action distribution.} \label{fig:general}
\end{figure*}
We evaluate R-GCN-GTG in eight tasks: five MinAtar games~\citep{young2019minatar}, a procedurally-generated LavaCrossing environment~\citep{gym_minigrid}, a box-world environment~\citep{zambaldi2018} requiring complex relational reasoning, and a symbolic variant of Read to Fight Monsters~\citep[RTFM,][]{zhong2019rtfm}, an environment that provides knowledge bases (KBs) describing environment dynamics that change in every episode.
On RTFM, we demonstrate that R-GCN-GTG can not only exploit the spatial information in a feature map, but in addition also relational information without modifying the neural architecture.
Our experiments show that R-GCN-GTG produces better policies than Convolutional Neural Networks (CNNs) or Neural Logic Machines~\citep[NLMs,][]{dong2018neural}, a state-of-the-art neural-symbolic model for relational reinforcement learning.

In summary, our main contributions are:
\begin{inparaenum}[\itshape i\upshape)]
\item we propose a principled approach for expressing relational inductive biases for neural networks in terms of relational graphs, % by building upon relational graphs of R-GCNs,
\item we introduce GTG to transform grid structures represented by a feature map into relational graphs that carrying spatial relational inductive biases,
\item we empirically demonstrate that in comparison to CNNs and NLMs, R-GCN-GTG generalizes better in both in- and out-of-distribution tasks in a diverse set of challenging procedurally-generated grid-world RL tasks, and finally 
\item we show that GTG is able to incorporate external knowledge, enabling R-GCN-GTG to jointly reason over spatial information and relational information about novel environment dynamics without any additional architectural modifications.
\end{inparaenum}
\section{Related Work}
\paragraph{Graph Neural Networks in RL}
To our knowledge, NerveNet~\cite{Wang2018} is the first work that used GNNs to represent an RL policy.
Their model follows a similar message-passing scheme as GCNs, maintaining only node feature vectors.
NerveNet has been bench-marked on MuJoCo environments, Snake and Centipede, and achieves better in-distribution performance and generalization than Multi-Layer Perceptrons (MLPs).
In these environments, NerveNet controls multi-joint robotic avatars.
State information and output actions are represented as a graph structure where each node corresponds to a movable part of the agent and local state and actions are attached to each node.
Their generalization tests cover size, disability transfer, and multi-task learning.
While achieving good systematic generalization performance, \citet{Wang2018} focus on the continuous control setting and abstract the graph structure from the morphological information of the avatar.
\citet{kurin2020body} propose Amorpheus, a transformer architecture for continuous control without relying on morphological information, outperforming NerveNet.
Our work focuses on a different class of tasks for which the input can be represented as a feature map.

\paragraph{Neuro-Logic Models for RL}
A separate branch of relational neuro-symbolic models builds directly upon first-order logic, for example, $\partial$ILP~\citep{Evans2018} and Neural Logic Machines (NLMs)~\citep{dong2018neural}.
Neural Logic Reinforcement Learning~\citep[NLRL,][]{pmlr-v97-jiang19a} applies a modified version of $\partial$ILP on a varied set of block world tasks and grid-world cliff-walking tasks, displaying robust generalization properties.
While $\partial$ILP has useful strong inductive biases that allow it to generalize well once it has learned a good policy, it suffers from poor scalability and proves difficult to train for more complex logical mappings.
Therefore, we use Neural Logic Machines~\citep[NLMs,][]{dong2018neural}, a more expressive and scalable neuro-symbolic architecture, as our baseline.
%Neural Logic Machines (NLMs)~\citep{dong2018neural} can be used to express DataLog clauses but the inductive bias is weaker than $\partial$ILP: the truth score of a head atom is represented by a linear combination of truth scores of the body atoms.
%
Besides supervised concept-learning tasks, NLMs also have been applied to simple reinforcement learning tasks~\citep{dong2018neural}.
In \citet{dong2018neural}, an NLM-based agent is trained to generalize to procedurally generated block world environments and algorithmic tasks. In these tasks, the NLM-based agent surpasses a Memory-Augmented Neural Networks baseline \cite{sukhbaatar2015end} both in terms of in-distribution generalization and out-of-distribution generalization to larger problem sizes.
%The simple tasks in \citet{dong2018neural} easily conform to a knowledge-base representation given their symbolic nature, which is not true for most RL tasks in general.
NLM treats relations as inputs and reasons about them in a soft, differentiable manner.
However, inductive biases for NLMs remain hard-coded in the architecture. %, which constrains the model to reason in a Datalog-like way.
In contrast, R-GCN uses learned relations to express relational inductive biases that constrain message passing.

\paragraph{Works Focusing on Symmetries.} 
Symmetries, especially equivariances, form an important class of relational inductive biases.
Some previous works~\cite{CohenW16,PolWHOW20} focus on how prior knowledge about symmetries can be incorporated into a model.
Group equivariant convolutional networks can represent arbitrary group symmetries, say, rotation and flipping~\cite{CohenW16}.
In RL, MDP Homomorphic Networks~\cite{PolWHOW20} express symmetries in the joint state-action space.
% Those methods concern a disjoint set of relational inductive bias that R-GCN-GTG can express.
For these methods, certain relational inductive biases outside of symmetries, like locality in CNNs, still must be hard-coded into the architecture.
R-GCN-GTG cannot express equivariance in the state-action space, but it can recover some symmetries like translation equivalence. However, it is unclear whether it can represent all group symmetries.

\section{Background}
\label{chapterlabel2}
\paragraph{Relational Graphs}
We define a relational graph as a labeled, directed multi-graph, denoted as $\mathcal{G}=(\mathcal{V}, \mathcal{E}, \mathcal{R})$, where $\mathcal{V}$ is the set of nodes (representing entities), $\mathcal{R}$ is the set of relation labels (representing relation types), $\mathcal{E} \subseteq \mathcal{V} \times \mathcal{R} \times \mathcal{V}$ is the set of relations (labeled directed edges).
Each relation is represented by a tuple $(a, r, b)$ with $a \in \mathcal{V}$, $r \in \mathcal{R}$, and $b \in \mathcal{V}$, and represents a relationship of type $r$ between the source entity $a$ and the target entity $b$ of the edge.
%
% have an example?
%

\paragraph{Relational Graph Convolutional Networks}
R-GCNs~\citep{schlichtkrull2018modeling} extend GNNs to model relational graphs.
R-GCNs represent a map from the set of feature vector nodes to a new set of feature vectors, conditioned on the relational graph $\mathcal{G}$ and the parameters $W_r$ attached to each relation label.
The update rule for a feature vector $\mb{x}_a$ of node $a$ is given by:
\begin{equation}
    \mb{x}^{\prime}_a := \sigma \left( \sum_{r\in \mathcal{R}} \sum_{b \in \mathcal{N}_a^r} \frac{1}{c_{a,r}} \mb{W}_r \mb{x}_b +\mb{W}_0 \mb{x}_a \right),
\end{equation}
where $\sigma$ is a non-linearity (such as the ReLu function), $\mathcal{N}_a^r$ denotes the neighboring nodes of $a$ under relation type $r$, $\mb{W}_r$ is the weight matrix associated with $r$, and $c_{a,r}$ is a normalization constant. In this work, we use $c_{a,r}=|\mathcal{N}_a^r|$.
R-GCNs were introduced to deal with graph structured data~\citep{schlichtkrull2018modeling}.
Our work presents a new perspective on R-GCNs: we view relational graphs as representing the connectivity and parameter sharing scheme for the model, thereby encoding a  prior relational inductive bias.

\section{Methodology}% with Relational Graphs}
In this section, we describe how relational graphs used by R-GCNs can be adopted to formalize two constraints commonly used in neural architecture design: sparse connectivity and parameter sharing.
We introduce GTG, a set of relation determination rules for representing spatial relational inductive biases. GTG strictly generalizes the inductive bias underlying convolutional layers.

Finally, we propose two ways of enabling R-GCNs to jointly reason with visual information restructured according to GTG and potentially additional, external relational knowledge.

\subsection{Expressing Relational Inductive Biases \\ Using Relational Graphs}
\label{sec:constraint}
In R-GCNs, message passing is explicitly directed by the relational graph rather than implicitly by the model architecture.
Removing the non-linearity and representing $W_0 \mb{x}_b$ as a self-loop edge~\footnote{All the self-loop edges share the same relation label, and term $W_0 \mb{x}_b$ is included the summation.}, we obtain the following simplified R-GCN update rule:
\begin{align}
    \mb{y}_a = \sum_{r\in \mathcal{R} \cup \{0\}} \sum_{b \in \mathcal{N}_a^r} \frac{1}{c_{a,r}} \mb{W}_r \mb{x}_b.
    \label{eq:simple}
\end{align}
The set of edges determines whether there is message passing between each pair of entities, thereby encoding the connectivity of the model.
The relationship labels indicate the specific pattern of parameters to be used by the message-passing functions.
By making use of different relational graphs, R-GCNs can represent many common neural architectures, including MLPs, CNNs, and DeepSets.
We provide a formal description of the neural architectures that R-GCN can represent in \cref{sec:block}.

To construct the relational graph, we make use of relation determination rules, each defined in the following form:
\begin{equation*}
    r(a,b) \leftarrow \text{condition},
\end{equation*}
\noindent where $r(a,b)$ is a relation from entity $a$ to $b$ with label $r$, and \emph{condition} is a logic statement.
If the condition holds true, relation label $r$ will be appended into $\mathcal{R}_{ab}$, the set of all relation labels of relations between entities $a$ and $b$.
The relation determination rules then express the relational inductive bias by controlling the sparsity and parameter sharing patterns of a feed forward neural network.
\subsection{Grid-to-Graph: Spatial Relational Inductive Biases} \label{sec:gtg}
We now introduce a set of relation determination rules that can be used to construct spatial relational inductive biases.
We start by replicating the relational inductive biases of CNNs, and then introduce new biases to address the limitation of CNNs.
\subsubsection{Local Directional Relations}
%
%The relation determination rules can carry relational inductive bias that will guide the message passing of R-GCNs so that further influence the performance of the models.
%
The number of possible definitions of spatial relationships between objects is very large, and it may not be feasible to enumerate each of them, let alone empirically evaluate them all.
We, therefore, start by mimicking the inductive biases encoded by CNNs, which have been shown to be effective in computer vision tasks and Deep Reinforcement Learning tasks with visual inputs~\citep{NIPS2012_4824,Mnih2015,silver2018general}. This provides us with a set of local directional relations.
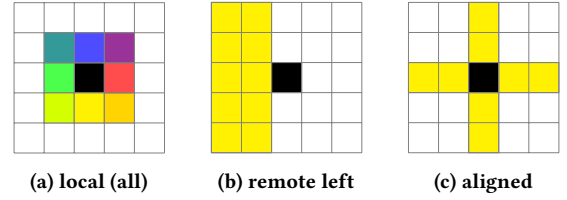
\begin{figure}[t]
    \begin{subfigure}[b]{0.3\columnwidth}
        \centering
        \begin{tikzpicture}
            \tikzmath{\d = 0.4;}
            \draw[step=0.4cm,gray,very thin] (0,0) grid (5*\d,5*\d);
            \filldraw[fill=black, draw=gray] (3*\d, 3*\d) rectangle (2*\d,2*\d);
            \filldraw[fill=green!70!white, draw=gray] (2*\d, 3*\d) rectangle (1*\d,2*\d);
           \filldraw[fill={rgb:green,1;yellow,5}, draw=gray] (2*\d, 2*\d) rectangle (1*\d,1*\d);
           \filldraw[fill=yellow, draw=gray] (3*\d, 2*\d) rectangle (2*\d,1*\d);
           \filldraw[fill={rgb:red,1;yellow,5}, draw=gray] (4*\d, 2*\d) rectangle (3*\d,1*\d);
            \filldraw[fill=red!70!white, draw=gray] (4*\d, 3*\d) rectangle (3*\d,2*\d);
           \filldraw[fill={rgb:red,2;blue,2;white,1}, draw=gray] (4*\d, 4*\d) rectangle (3*\d,3*\d);
            \filldraw[fill=blue!70!white, draw=gray] (3*\d, 4*\d) rectangle (2*\d,3*\d);
           \filldraw[fill={rgb:green,2;blue,2;white,1}, draw=gray] (2*\d, 4*\d) rectangle (1*\d,3*\d);
        \end{tikzpicture}
        \caption{local (all)}
    \end{subfigure}
    \begin{subfigure}[b]{0.3\columnwidth}
        \centering
        \begin{tikzpicture}
            \tikzmath{\d = 0.4;}
            \filldraw[fill=black, draw=gray] (3*\d, 3*\d) rectangle (2*\d,2*\d);
           \filldraw[fill=yellow, draw=gray] (0*\d, 0*\d) rectangle (2*\d,5*\d);
            \draw[step=0.4cm,gray,very thin] (0,0) grid (5*\d,5*\d);
        \end{tikzpicture}
        \caption{remote left}
    \end{subfigure}
    \begin{subfigure}[b]{0.3\columnwidth}
        \centering
        \begin{tikzpicture}
            \tikzmath{\d = 0.4;}
            \filldraw[fill=yellow, draw=gray] (2*\d, 0*\d) rectangle (3*\d,5*\d);
            \filldraw[fill=yellow, draw=gray] (0*\d, 2*\d) rectangle (5*\d,3*\d);
            \filldraw[fill=black, draw=gray] (3*\d, 3*\d) rectangle (2*\d,2*\d);
            \draw[step=0.4cm,gray,very thin] (0,0) grid (5*\d,5*\d);
        \end{tikzpicture}
        \caption{aligned}
    \end{subfigure}
    \caption{Three subsets of GTG relation determination rules: the black tile is the target node, and the other colored tiles are source nodes. Tiles with the same relationship to the target node share the same color.} \label{fig:i2k}
\end{figure}
Each local directional relation specifies the relative position of two adjacent entities.
A graphical illustration is shown in \cref{fig:i2k} (a), where a selected target node is painted black and source nodes are painted with different colors, each corresponding to different relation labels.
For clarity, we picked a single node as the target node, though this may not be generally the case.
Visualizing all the local directional relations would result in a mesh of edges connecting all nodes to each other.
Consider two entities $a$ and $b$, and their coordinates $x_a$, $y_a$, $x_b$, $y_b$.
The determination rules for local directional relations are as follows:
\begin{equation*}
\begin{aligned}
    \text{rightAdj}(a,b) &\leftarrow (x_a=x_b+1) \wedge (y_a=y_b), \\
    \text{leftAdj}(a,b) &\leftarrow (x_a=x_b-1) \wedge (y_a=y_b), \\
    \text{topAdj}(a,b) &\leftarrow (y_a=y_b+1) \wedge (x_a=x_b), \\
    \text{bottomAdj}(a,b) &\leftarrow (y_a=y_b-1) \wedge (x_a=x_b),
\end{aligned}
\end{equation*}
\begin{equation*}
\begin{aligned}
    \text{topRightAdj}(a,b) &\leftarrow (x_a=x_b+1) \wedge (y_a=y_b+1), \\
    \text{topLeftAdj}(a,b) &\leftarrow (x_a=x_b-1) \wedge (y_a=y_b+1), \\
    \text{bottomRightAdj}(a,b) &\leftarrow (x_a=x_b+1) \wedge (y_a=y_b-1), \\
    \text{bottomLeftAdj}(a,b) &\leftarrow (x_a=x_b-1) \wedge (y_a=y_b-1). 
\end{aligned}
\end{equation*}
By only applying these local directional relations, the computation of the associated R-GCN model would be equivalent to that of a convolutional layer with a $3 \times 3$ kernel (up to a normalization constant).

\subsubsection{Remote Directional Relations}
One limitation of convolutional layers is the difficulty of message passing among remote entities.
In order to pass information to another node N blocks away, N layers are needed for a CNN with strides length of 1.
This problem can be alleviated with larger strides, pooling layers, or dilated convolutions~\citep{yu2015multi,goodfellow2016deep}, but the model will still require a large number of layers.
For less deep CNNs, such as the baseline model used in this work, long-distance message passing is accomplished using dense layers following the convolution layers, as there is no message passing between distant entities.
However, dense layers exhibit only a weak relational inductive bias, which can hurt generalization performance. With only local directional relations, R-GCNs inherit the same long distance message passing problem as convolution layers.
We, therefore, introduce remote directional relations, which capture the notion of relative positions between objects. 
We visualize one such remote directional relation, \emph{left}, in \cref{fig:i2k} (b). We express remote directional relations using the following rules:
\begin{equation*}
\begin{aligned}
    \text{right}(a,b) &\leftarrow x_a>x_b, \\
    \text{left}(a,b) &\leftarrow x_a<x_b, \\
    \text{top}(a,b) &\leftarrow y_a>y_b, \\
    \text{bottom}(a,b) &\leftarrow y_a<y_b.
\end{aligned}
\end{equation*}

\subsubsection{Alignment and Adjacency Relations} 
\label{sec:align}
Besides these directional relations, we also add two auxiliary relations, \emph{aligned} and \emph{adjacent}:
\begin{equation*}
\begin{aligned}
\text{aligned}(a,b) &\leftarrow \left(x_a=x_b\right) \lor \left(y_a=y_b\right),\\
\text{adjacent}(a,b) &\leftarrow \left(|x_a-x_b| \le 1\right) \land \left(|y_a-y_b| \le 1\right).
\end{aligned}
\end{equation*}
\emph{Aligned} relations indicate if two entities are on the same horizontal or vertical line, visualized in \cref{fig:i2k} (c).
\emph{Adjacent} relations indicate whether two objects are adjacent to each other, which, unlike local directional relations, carry no directional information. 

\subsection{Jointly Reasoning with External Knowledge}
GTG expresses spatial relational inductive biases in the form of a relational graph.
As R-GCN was originally designed to reason over knowledge graphs, it may be tempting to let R-GCN jointly reason over spatial inputs and a task-relevant external relational knowledge graph by simply merging some graph representation of each without further architecture changes. We introduce two ways of incorporating external relational knowledge: one-hop relations between physical entities and grounding relations with a knowledge graph.
Examples of these two approaches applied to RTFM can be found in \cref{sec:rtfm} and \cref{fig:rtfm}.

\subsubsection{Relations between Physical Entities}
We refer to each cell in the feature map a physical entity. We can then straightforwardly introduce relational knowledge by adding relations between physical entities.
However, this limits the knowledge that can be expressed, as this approach cannot represent more abstract knowledge that describes relations between concepts rather than specific entities, e.g., a shinning weapon can kill fire monsters.

\subsubsection{Working with External Knowledge Bases}
For enabling the inclusion of external knowledge in our model, we maintain two sets of entities: conceptual entities, which exist in the knowledge base (e.g. the class of an object) and physical entities, which exist in the environment (e.g. a specific object in the environment, such as a monster).
We can then link the two graphs corresponding to these two sets of entities with grounding relations so that information can flow between them.

\begin{figure}[t]
    \centering
    \includegraphics[width=\linewidth]{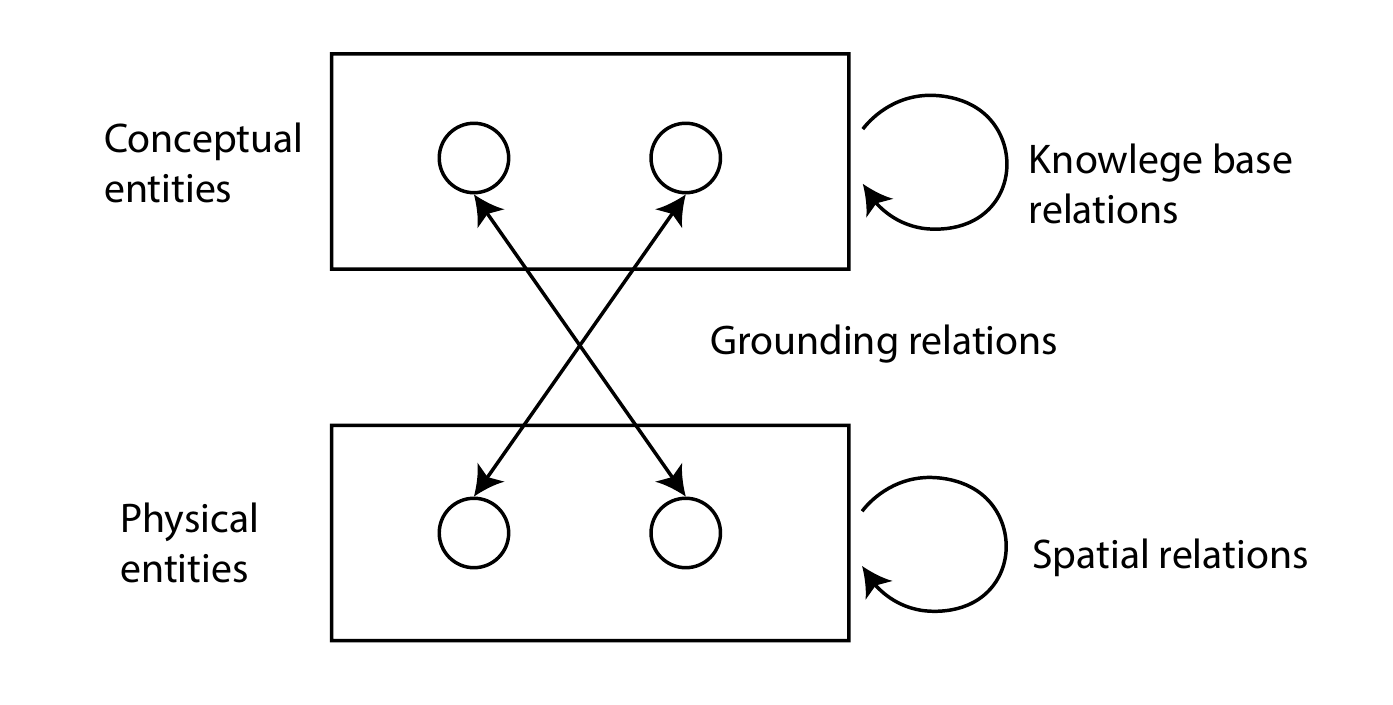}
    \caption{Reasoning with an external knowledge base}
    \label{fig:grounding}
\end{figure}

A graphical illustration of this approach is presented in \cref{fig:grounding}.
The grounding relations assign conceptual counterparts to the physical entities. In this work, these relations are handcrafted. 
\section{Experimental Setting}
\subsection{MinAtar}

MinAtar~\citep{Young2019} is a collection of miniature Atari games.
Each observation is a $10 \times 10$ feature map, where each cell represents a single object in the game. Unlike conventional Atari games, the environments in MinAtar games are stochastic -- for instance in Breakout, the ball starts in a random position.
MinAtar environments also set a 10\% sticky action~\citep{hausknecht2015impact} probability by default. Sticky actions force the agent to take the action taken in the last step. 
We enforce a 5,000 steps limit, as some agents can play indefinitely on some of the MinAtar games.
%
%

%
\iffalse
%
\begin{figure}[t]
\centering
     \begin{subfigure}[b]{0.19\textwidth}
     \includegraphics[width=\textwidth]{img/annotated_seaquest.png}
     \caption{seaquest}
     \end{subfigure}
     \begin{subfigure}[b]{0.19\textwidth}
     \includegraphics[width=\textwidth]{img/annotated_asterix.png}
     \caption{asterix}
     \end{subfigure}
     \begin{subfigure}[b]{0.19\textwidth}
     \includegraphics[width=\textwidth]{img/annotated_freeway.png}
     \caption{freeway}
     \end{subfigure}
     \begin{subfigure}[b]{0.19\textwidth}
     \includegraphics[width=\textwidth]{img/annotated_breakout.png}
     \caption{breakout}
     \end{subfigure}
     \begin{subfigure}[b]{0.19\textwidth}
     \includegraphics[width=\textwidth]{img/annotated_space_invaders.png}
     \caption{space invaders}
     \end{subfigure}
     \caption{The annotated game state of the MinAtar games, produced by the MinAtar paper \cite{young2019minatar}.}
     \label{fig:minatar}
\end{figure}
%
\fi
%

%
\subsection{Box-World}
Box-world is a grid-world navigation game introduced by \citet{zambaldi2018deep}.
To solve the game, the agent must collect the gem using the correct key.
However, the key is in a locked box, which needs to be unlocked with a separate key.
There are also distractor branches which will consume the current key and produce a key that cannot unlock the gem box.
As the game is combinatorially complex, the chance of hitting the correct solution by random walk is low.
\citet{zambaldi2018deep} demonstrated that their RL models required between $2\times 10^8$ to $14\times 10^8$ steps to converge in this environment.
Due to limited computational resources, we only train models for $10^7$ steps in each environment and further reduce the difficulty of the Box-World environment:
\begin{inparaenum}[\itshape 1\upshape)]
\item the field size is reduced to $10\times 10$,
\item the number of distractor branches is set to 1,
\item the length of distractor branches is set to 1,
\item the goal length is set to 2.
\end{inparaenum}
Although we reduced the difficulty, we still preserve all core elements of the Box-World environment.
We expect this simplified experimental setup to still allow us to compare the relational reasoning capabilities of different models while reducing the total steps needed for training to convergence.
\subsection{LavaCrossing}
LavaCrossing is a standard environment from MiniGrid~\citep[Minimalistic Gridworld,][]{gym_minigrid}.
The agent must navigate to the goal position without falling into the lava river.
The game is procedurally generated, and there are 3 difficulty levels available for each map size, making this is an ideal RL environment to test combinatorial and out-of-distribution generalization of learned policies.
MiniGrid environments are by default partially-observable, but we configure our instances to be full-observable.
Also, by default, an agent can turn left, turn right, and move forward, which requires the agent to know its direction when navigating to a particular position. We adjust the action space, making the agent able to move in all four directions without turning left or right.
%
\iffalse
%
\begin{figure}[t]
\centering
     \begin{subfigure}[b]{0.32\textwidth}
     \includegraphics[width=\textwidth]{img/LavaCrossingS9N1.png}
     \caption{Level 1}
     \end{subfigure}
     \begin{subfigure}[b]{0.32\textwidth}
     \includegraphics[width=\textwidth]{img/LavaCrossingS9N2.png}
     \caption{Level 2}
     \end{subfigure}
     \begin{subfigure}[b]{0.32\textwidth}
     \includegraphics[width=\textwidth]{img/LavaCrossingS9N3.png}
     \caption{Level 3}
     \end{subfigure}
     \caption{Initial states in LavaCrossing for various difficulty levels. Initial states per episodes will differ, as levels are procedurally generated.}
     \label{fig:lavacrossing}
\end{figure}
%
\fi
%
The number of lava rivers generated equals the level number.
To test the out-of-distribution generalization, we train the agent on difficulty level 2 and test on difficulty levels 1 and 3.

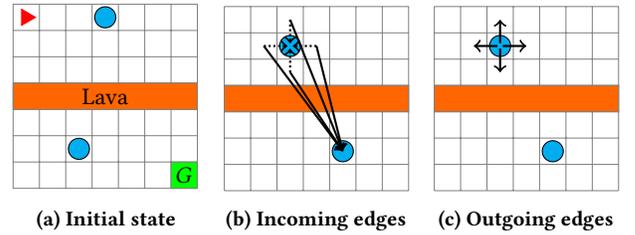
\begin{figure}[t]
    \centering
    \begin{subfigure}[b]{0.32\columnwidth}
        \centering
        \begin{tikzpicture}
            \tikzmath{\d = 0.35;}
            \draw[step=\d,gray,very thin] (0,0) grid (7*\d,7*\d);
            \filldraw[fill=red!60!yellow, draw=gray] (0,3*\d) rectangle (7*\d,4*\d);
            \filldraw[cyan, draw=black] (2.5*\d, 1.5*\d) circle (0.4*\d);
            \filldraw[cyan, draw=black] (3.5*\d, 6.5*\d) circle (0.4*\d);
            \node[](lava) at (3.5*\d,3.5*\d){Lava};
            \filldraw[fill=green, draw=gray] (6*\d, 0) rectangle (7*\d,\d);
            \node[](G) at (6.5*\d,0.5*\d){$G$};
            \node[mark size=10*\d,color=red,rotate=30] at (0.5*\d,6.5*\d) {\pgfuseplotmark{triangle*}};
        \end{tikzpicture}
        \caption{Initial state}
    \end{subfigure}
    \begin{subfigure}[b]{0.32\columnwidth}
        \centering
        \begin{tikzpicture}
            \tikzmath{\d = 0.35;}
            \draw[step=\d,gray,very thin] (0,0) grid (7*\d,7*\d);
            \filldraw[fill=red!60!yellow, draw=gray] (0,3*\d) rectangle (7*\d,4*\d);
            \filldraw[cyan, draw=black] (4.5*\d, 1.5*\d) circle (0.4*\d);
            \filldraw[cyan, draw=black] (2.5*\d, 5.5*\d) circle (0.4*\d);
            \draw[densely dotted,thick,->] (1.5*\d,5.5*\d) -- (2.4*\d, 5.5*\d);
            \draw[densely dotted,thick,->] (3.5*\d,5.5*\d) -- (2.6*\d, 5.5*\d);
            \draw[densely dotted,thick,->] (2.5*\d,4.5*\d) -- (2.5*\d, 5.4*\d);
            \draw[densely dotted,thick,->] (2.5*\d,6.5*\d) -- (2.5*\d, 5.6*\d);
            \draw[thick,->] (1.5*\d,5.5*\d) -- (4.5*\d, 1.5*\d);
            \draw[thick,->] (3.5*\d,5.5*\d) -- (4.5*\d, 1.5*\d);
            \draw[thick,->] (2.5*\d,4.5*\d) -- (4.5*\d, 1.5*\d);
            \draw[thick,->] (2.5*\d,6.5*\d) -- (4.5*\d, 1.5*\d);
        \end{tikzpicture}
        \caption{Incoming edges}
    \end{subfigure}
    \begin{subfigure}[b]{0.32\columnwidth}
        \centering
        \begin{tikzpicture}
            \tikzmath{\d = 0.35;}
            \draw[step=\d,gray,very thin] (0,0) grid (7*\d,7*\d);
            \filldraw[fill=red!60!yellow, draw=gray] (0,3*\d) rectangle (7*\d,4*\d);
            \filldraw[cyan, draw=black] (4.5*\d, 1.5*\d) circle (0.4*\d);
            \filldraw[cyan, draw=black] (2.5*\d, 5.5*\d) circle (0.4*\d);
            \draw[thick,->](2.4*\d, 5.5*\d) -- (1.5*\d,5.5*\d);
            \draw[thick,->] (2.6*\d, 5.5*\d) --(3.5*\d,5.5*\d);
            \draw[thick,->] (2.5*\d, 5.4*\d) --(2.5*\d,4.5*\d);
            \draw[thick,->] (2.5*\d, 5.6*\d) --(2.5*\d,6.5*\d);
        \end{tikzpicture}
        \caption{Outgoing edges}
    \end{subfigure}
    \caption{Portal-LavaCrossing tasks. (a) One possible initial state, where blue circles are portals, the red triangle is the agent, and the green block is the goal position. (b) How incoming edges to the portal are attached to the paired portal. Dashed arrows represent the original edges and solid arrows, new edges. (c) Outgoing edges from the portal remain the same.} \label{fig:portal}
\end{figure}

We also design a Portal-LavaCrossing task, illustrated in \mbox{\cref{fig:portal} (a)}, to test whether R-GCN-GTG can generalize to non-Euclidean spaces without retraining.
After training on difficulty level 2, we transfer the agent to Portal-LavaCrossing where there are no gaps in the lava river.
For each side of the lava river, a teleportation portal is placed in a random position: when the agent moves into the portal, it is then placed on the other side of the lava river, and this is the only way to cross it.
As no such portals exist in training levels, the agent  must be able to generalize to a new environment with non-Euclidean space, leveraging novel test-time spatial relations, or resort to moving around randomly until stepping into the portal.
In this task, the CNN baseline is not aware of the portal at all, and can only reach the goal if it moves into the portal by chance.
For R-GCNs and NLMs, we append new spatial relationships between portals and other grids:
incoming relations to one portal all connect to the paired portal on the other side of the lava river, and
outgoing relations from one portal are kept the same.
These relations are shown in \cref{fig:portal} (b) and (c).

\subsection{Read to Fight Monsters} \label{sec:rtfm}
%\tim{this description needs to be shortened considerably (just refer to the original paper) and focus on what's important of the environment for our purposes of evaluation GTG and relational RL}
Read to Fight Monsters~\citep{zhong2019rtfm} is a grid world game, where each level includes a text document providing information about per-episode game dynamics.
Each map contains two monsters and two weapons, each randomly generated and positioned.
Each weapon has a modifier and each monster has an element property.
The agent must defeat the monster of a specific element, which can only be defeated with weapons with a specific modifier.
The relations describing which modifiers defeat which elements are procedurally generated at the start of each episode and described by the document.
Furthermore, each monster belongs to a team. The text document also describes which team must be defeated.
Without the text document, the agent can only pick a weapon and attack an arbitrary monster, which leads to an overall win probability of less than $50\%$.
We construct a knowledge base which contains the same information as the original RTFM document based on the grammatical rules that RTFM uses to generate the text document. 

\begin{figure}[t]
    \centering
    \begin{subfigure}[b]{0.35\columnwidth}
        \centering
        \begin{tikzpicture}
            \tikzmath{\d = 0.4;}
            \draw[step=0.4cm,gray,very thin] (0,0) grid (6*\d,6*\d);
            \filldraw[fill=green, draw=gray] (5*\d, 0) rectangle (6*\d,\d);
            \node[]() at (5.5*\d,0.5*\d){W};
            \filldraw[fill=green, draw=gray] (1*\d, 5*\d) rectangle (2*\d,6*\d);
            \node[]() at (1.5*\d,5.5*\d){W};
            \filldraw[fill=cyan, draw=red, line width=0.5mm] (2*\d, 2*\d) rectangle (3*\d,3*\d);
            \node[]() at (2.5*\d,2.5*\d){M};
            \filldraw[fill=cyan, draw=gray] (4*\d, 4*\d) rectangle (5*\d,5*\d);
            \node[]() at (4.5*\d,4.5*\d){M};
            \node[mark size=5pt,color=red,rotate=30] at (0.5*\d,4.5*\d) {\pgfuseplotmark{triangle*}};
            \draw[line width=0.5mm,->]  (5.2*\d,0.8*\d) -- (2.8*\d,2.2*\d);
            \draw[line width=0.5mm,->]  (1.8*\d,5.2*\d) -- (4.2*\d,4.8*\d);
        \end{tikzpicture}
        \caption{RTFM-onehop-KB}
    \end{subfigure}
    \begin{subfigure}[b]{0.55\columnwidth}
        \centering
        \begin{tikzpicture}
            \tikzmath{\d = 0.4;}
            \tikzmath{\wx = 7; \wy = 0; \mx = 3;}
            \filldraw[fill=green, draw=gray] (\wx*\d, \wy*\d) rectangle ({(\wx+1)*\d},{(\wy+1)*\d});
            \node[](W) at ({(\wx+0.5)*\d},{(\wy+0.5)*\d}){W};
            \filldraw[fill=cyan, draw=gray] (\mx*\d, \wy*\d) rectangle ({(\mx+1)*\d},{(\wy+1)*\d});
            \node[](M) at ({(\mx+0.5)*\d},{(\wy+0.5)*\d}){M};
            \tikzmath{\ix = -1; \iy = 2; \idx=2.75; \idy=3.5;}
            \draw[color=gray] (\ix*\d, \iy*\d) rectangle ({(\ix+\idx)*\d},{(\iy+\idy)*\d});
            \node[]() at ({(\ix+\idx*0.5)*\d},{(\iy+\idy+0.5)*\d}) {\footnotesize teams};
            \node[rectangle, draw=red, line width=0.5mm](T) at ({(\ix+\idx*0.5)*\d},{(\iy+\idy-2)*\d}) {\footnotesize alliance};
            \node[]() at ({(\ix+\idx*0.5)*\d},{(\iy+\idy-1)*\d}) {...};
            \draw[color=gray] (\wx*\d, \iy*\d) rectangle ({(\wx+\idx)*\d},{(\iy+\idy)*\d});
            \node[]() at ({(\wx+\idx*0.5)*\d},{(\iy+\idy+0.5)*\d}) {\footnotesize modifiers};
            \node[](O) at ({(\wx+\idx*0.5)*\d},{(\iy+\idy-2)*\d}) {\footnotesize master};
            \node[]() at ({(\wx+\idx*0.5)*\d},{(\iy+\idy-1)*\d}) {...};
            \draw[color=gray] (\mx*\d, \iy*\d) rectangle ({(\mx+\idx)*\d},{(\iy+\idy)*\d});
            \node[]() at ({(\mx+\idx*0.5)*\d},{(\iy+\idy+0.5)*\d}) {\footnotesize elements};
            \node[](E) at ({(\mx+\idx*0.5)*\d},{(\iy+\idy-2)*\d}) {\footnotesize fire};
            \node[]() at ({(\mx+\idx*0.5)*\d},{(\iy+\idy-1)*\d}) {...};
            \draw [->,line width=0.5mm] (T) edge (M);
            \draw [->,line width=0.5mm] (O) edge (W);
            \draw [->,line width=0.5mm] (E) edge (M);
            \draw [->,line width=0.5mm] (O) edge (E);
        \end{tikzpicture}
        \caption{RTFM-KB}
    \end{subfigure}
    \caption{RTFM tasks with KB encodings of the text document. The red triangle represents the agent; the blue block, the monster; and the green block, the weapon. Figure (a) shows the one-hop reasoning version of KB encoding, where arrows indicate beat relations, and the red frame indicates the target monster.
    Figure (b) illustrates the multi-hop reasoning version.  Arrows indicate a relation between two entities.
    The red frame indicates the opponent team.
    Relation labels are not represented explicitly in the graph.} \label{fig:rtfm}
\end{figure}
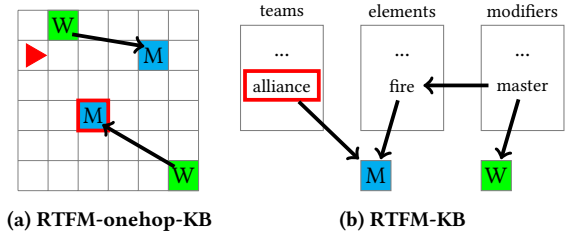

We test two approaches of introducing external knowledge to RTFM, shown in \cref{fig:rtfm}.
The easier physical-entities-only approach uses two relation labels \textit{target} and \textit{beat}, ignoring the concept of modifiers, elements, and teams.
\textit{target}($a$) is a unary atom indicating \textit{monster} $a$ is the one the agent must defeat and is appended to feature vectors.
The relation \textit{beat}($a$,$b$) (binary atom) means \textit{weapon} $a$ defeats \textit{monster} $b$.
If the agent carries the weapon, \textit{entity} $a$ corresponds to the agent itself.
A more complex approach introduces conceptual entities and uses multi-hop reasoning to solve the problem.
Along with the physical objects in the environment, this approach considers the conceptual entities of teams, modifiers and elements.
This approach introduces additional grounding relations: The relation \textit{assign}($a$,$b$) assigns \textit{modifier} $a$ to \textit{weapon} $b$ or \textit{element} $a$ to \textit{monster} $b$.
The relation \textit{belong}($a$,$b$) indicates that the \textit{monster} $a$ belongs to \textit{team} $b$.
The relation \textit{beat}($a$,$b$) states that \textit{modifier} $a$ defeats monsters of \textit{element} $b$.
The relation \textit{target}($a$) means that the agent must defeat \textit{team} $a$.
Finally, \textit{hold}($a$) indicates the agent currently holds a weapon with \textit{modifier} $a$.

\subsection{Architecture Overview}
In this subsection, we describe how GTG and R-GCN can be incorporated into an RL policy.
First, the state of the environment is rendered as a feature map.
Specifically, each tile in a grid world is represented as a binary-valued feature vector $\mb x$.
These feature vectors $\mathcal{X}$ are attached to nodes and GTG generates edges between these nodes, forming a relational graph $\mathcal{G}$ that represents particular relational inductive biases.
If required, extra knowledge about game dynamics expressed as a knowledge base can be merged into this multigraph.
Subsequently, the R-GCN acts on this multi-graph and associated feature vectors.
After processing using R-GCN, we apply a feature-wise max-pooling to all node feature vectors. 
The outputs are then fed to dense layers that output per-action logits.
The graphical illustration of the whole process can be found in \cref{fig:general}.
%\tim{this needs to be formalized, and ideally also visualized with an overview figure}
% The action logits will eventually be fed to a softmax function so that we get the probabilities for discrete actions.
The probability of actions can thus be written as:
\begin{equation}
    P(\mb{a}|\mathcal{X})=\text{softmax}(\text{MLP}(\text{maxpool}(g(\mathcal{X},\mathcal{G};\pmb{\theta}_1));\pmb{\theta}_2)),
\end{equation}
where $g$ is the stack of R-GCN layers and $\pmb \theta$s are neural network parameters.

A separate head performs value estimation:
\begin{equation}
    \hat v(\mathcal{X})=\text{MLP}(\text{maxpool}(g(\mathcal{X},\mathcal{G};\pmb{\theta}_1));\pmb{\theta}_3).
\end{equation}

%For NLM, we apply several steps of forward logic chaining.
%To get the propositional outputs, we append a dense (output) layer taking the output evaluations of nullary atoms as input and outputting the action logits.\tim{this needs to be formalized, and ideally also visualized with an overview figure; also not much reason to introduce NLM formally in the background if we don't make use of the notation introduced there at this point}

During training, the agent samples actions according to this resultant action distribution.
During testing, we take the action with maximum probability.
In our experiments, we make use of IMPALA \cite{espeholt2018impala}, a policy-gradient algorithm to train our RL models.
We based our IMPALA implementation on TorchBeast\cite{torchbeast2019} and our R-GCN implementation, on Pytorch-Geometric\cite{Fey/Lenssen/2019}.
Our implementation is available at \url{https://github.com/ZhengyaoJiang/GTG}.
\section{Results and Discussion}
In this section, we report our empirical results comparing R-GCN to baseline methods regarding in-distribution performance, out-of-distribution combinatorial generalization, and the ability to incorporate external knowledge.
We also report ablations probing the effectiveness of different relation determination rules and components of R-GCN-GTG.

\subsection{In-Distribution Generalization}
\cref{fig:indistribution} shows the in-distribution performance (training curve) of CNNs and relational models on MinAtar and LavaCrossing tasks. 
For each model, we run 5 trials, each with different random seeds.
The thin, opaque lines in the plot represent training curves corresponding to each run, and the bolded lines represent mean episodic return averaged over all five runs.
Here, NLM and R-GCN use GTG with all three classes relationships, namely, local directional, remote directional, and auxiliary relations.
We can see that R-GCN-GTG models consistently perform either better or on-par with CNNs and NLM-GTGs across all eight environments.
For Asterix, Seaquest, Box-World and Breakout, R-GCN-GTG outperforms CNNs by a significant margin.
As the CNN baseline in the RTFM environment is unable to access information in the knowledge base, it acts as an informative baseline, for which only visual information is available.
NLM-GTGs also achieve good performance on Seaquest and Breakout, but they are inferior to CNNs on other MinAtar tasks.
We also mark the best performance of original MinAtar baselines \cite{young2019minatar} as a red horizontal dashed line.
It is worth noting that we use a deeper network than these baselines, which include a deep Q-network \cite{mnih2015human} and actor-critic with eligibility traces \cite{degris2012model,SuttonBook}, trained on twice as many steps.
Thus, our CNN agent acts as a much stronger baseline than the original MinAtar models.

There are two potential reasons for the large performance gain of R-GCN-GTG over CNNs.
Firstly, R-GCN-GTG does not make use of the absolute position of objects, but instead only taking into account the relative positions between objects.
In contrast, CNNs employ dense layers to reason globally. Further, these dense layers have a weak relational inductive bias which can negatively impact sample efficiency and generalization.
% Such an ability to generalize is important for the in-distribution performance in stochastic or procedurally generated environments, as it is difficult to exhaustively memorise all optimal trajectories.
Secondly, GTG provides more flexible message passing than conventional CNN layers in terms of long-range dependencies.
%In CNNs, such long-range dependencies can only be considered with a large receptive field through stacking multiple CNN layers or using large strides. 
%In contrast, GTG expresses long-range dependencies directly via remote directional relations.
We further study the roles of various relation determination rules and max-pooling after convolutions in our ablation studies.

\newcommand\x{0.49}
\begin{figure*}[t]
\centering
     \begin{subfigure}[b]{\x\columnwidth}
     \includegraphics[width=\columnwidth]{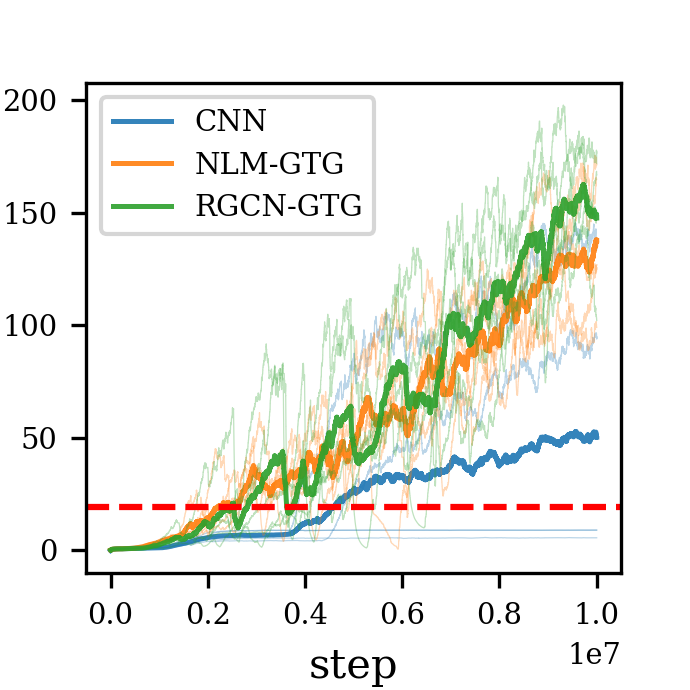}
     \caption{Seaquest}
     \end{subfigure}
     \begin{subfigure}[b]{\x\columnwidth}
     \includegraphics[width=\columnwidth]{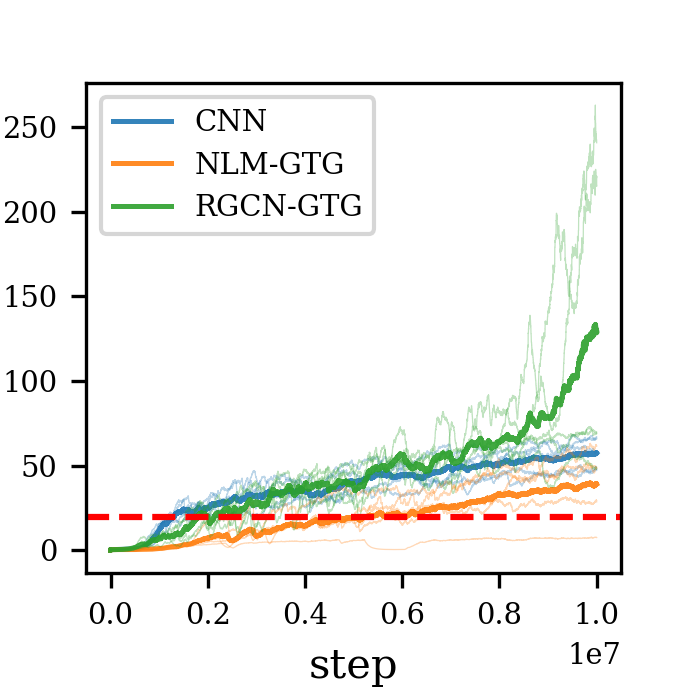}
     \caption{Asterix}
     \end{subfigure}
     \begin{subfigure}[b]{\x\columnwidth}
     \includegraphics[width=\columnwidth]{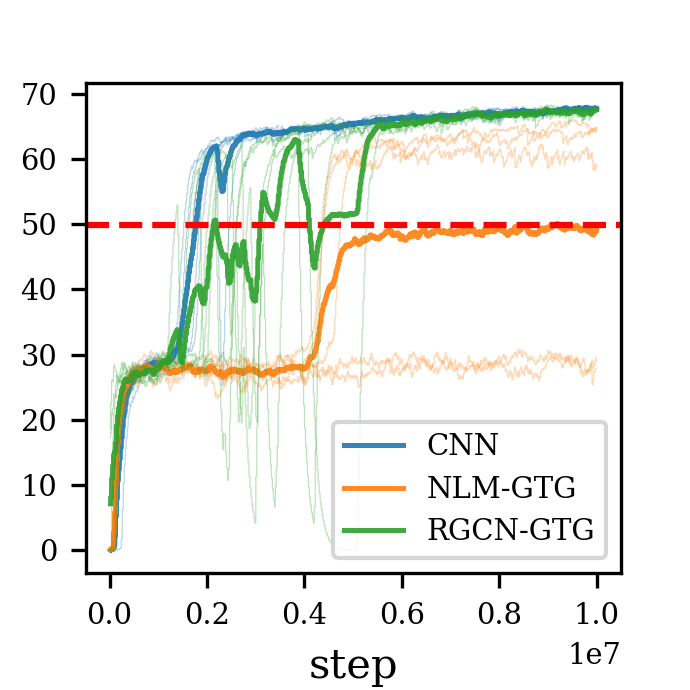}
     \caption{Freeway}
     \end{subfigure}
     \begin{subfigure}[b]{\x\columnwidth}
     \includegraphics[width=\columnwidth]{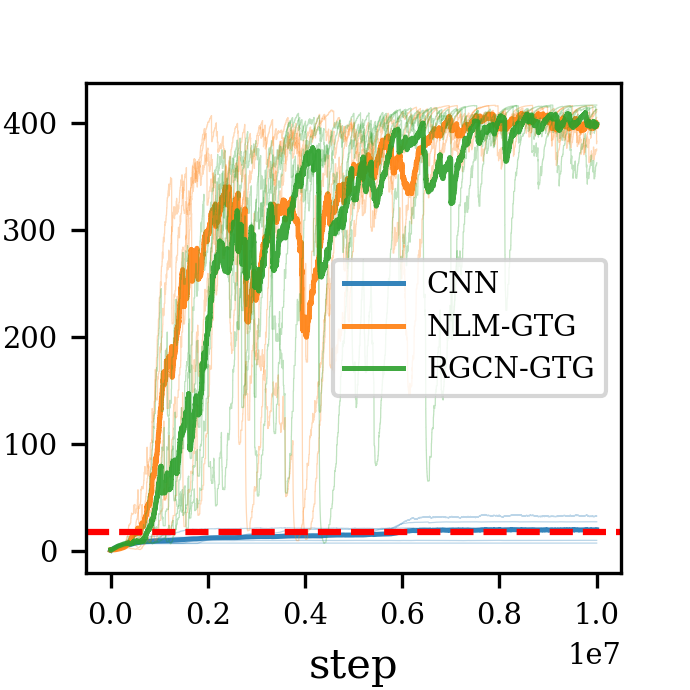}
     \caption{Breakout}
     \end{subfigure}
     \begin{subfigure}[b]{\x\columnwidth}
     \includegraphics[width=\columnwidth]{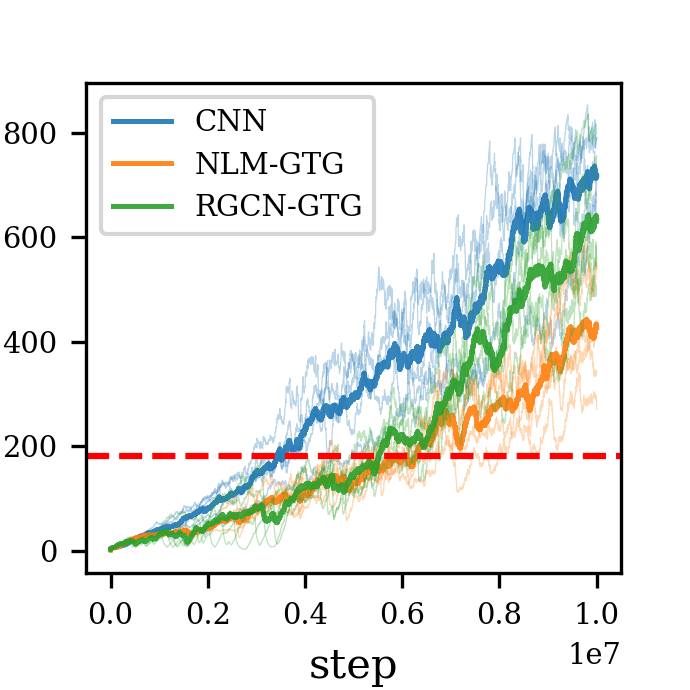}
     \caption{Space Invaders}
     \end{subfigure}
     \begin{subfigure}[b]{\x\columnwidth}
     \includegraphics[width=\columnwidth]{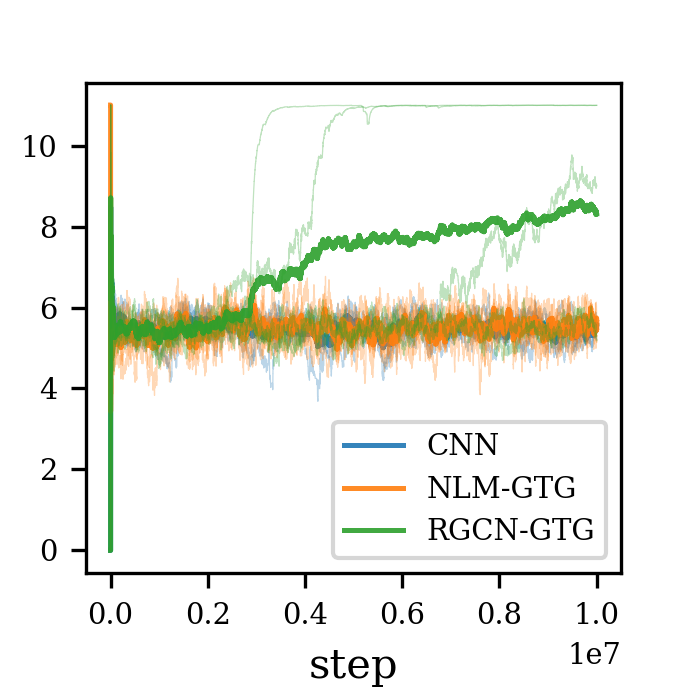}
     \caption{Box-World}
     \end{subfigure}
     \begin{subfigure}[b]{\x\columnwidth}
     \includegraphics[width=\columnwidth]{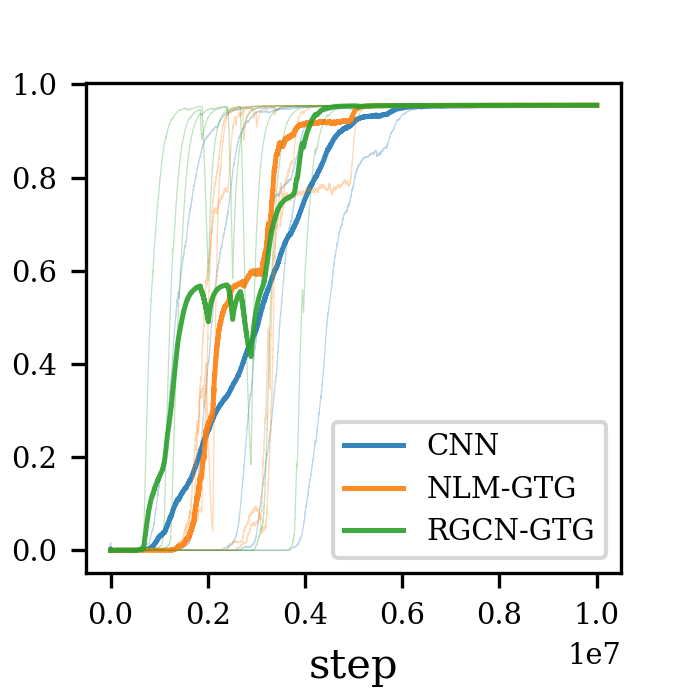}
     \caption{LavaCrossing}
     \end{subfigure}
     \begin{subfigure}[b]{\x\columnwidth}
     \includegraphics[width=\columnwidth]{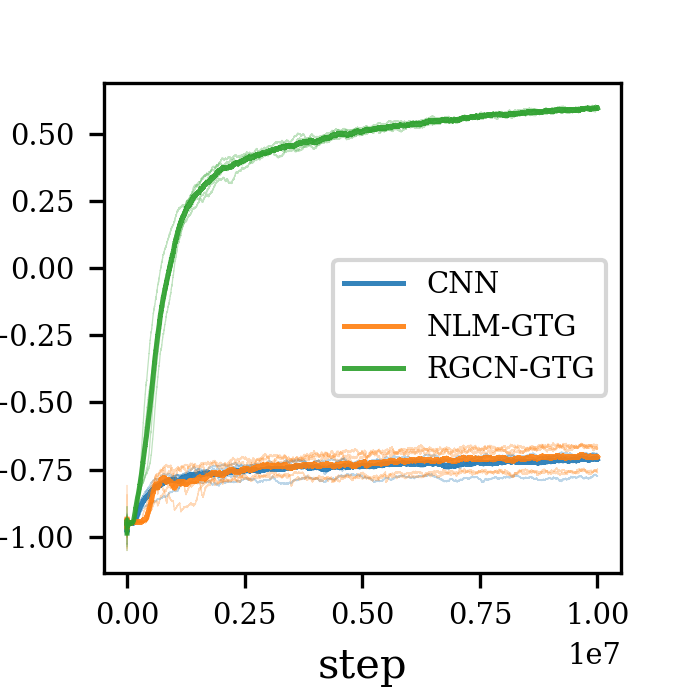}
     \caption{RTFM-KB}
     \end{subfigure}
     \caption{Training curves of CNN and relational models. The opaque lines represent the returns of individual runs, while the bolded lines, the average of 5 runs. Red dashed lines mark the final performance of the best model reported by the original MinAtar baselines.}
     \label{fig:indistribution}
\end{figure*}

\begin{figure}[t]
\centering
     \begin{subfigure}[b]{0.48\columnwidth}
     \includegraphics[width=\columnwidth]{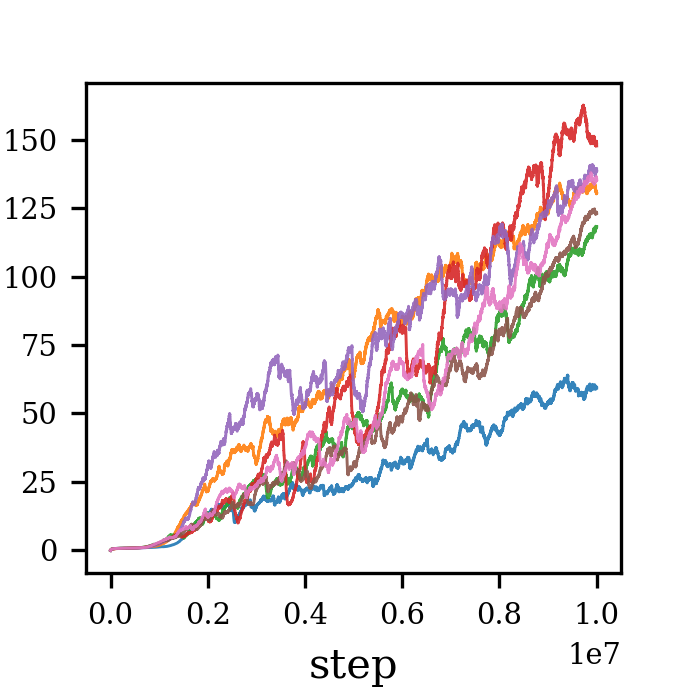}
     \caption{Seaquest}
     \end{subfigure}
     \begin{subfigure}[b]{0.48\columnwidth}
     \includegraphics[width=\columnwidth]{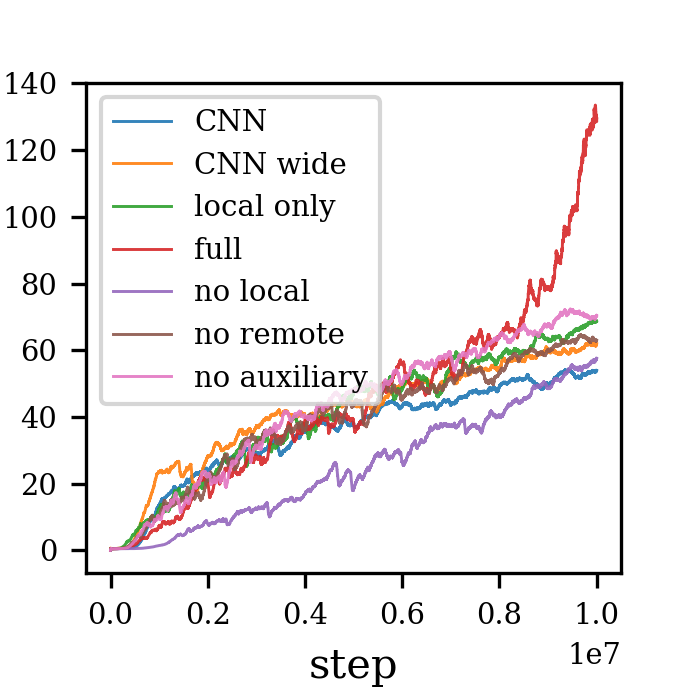}
     \caption{Asterix}
     \end{subfigure}
     \begin{subfigure}[b]{0.48\columnwidth}
     \includegraphics[width=\columnwidth]{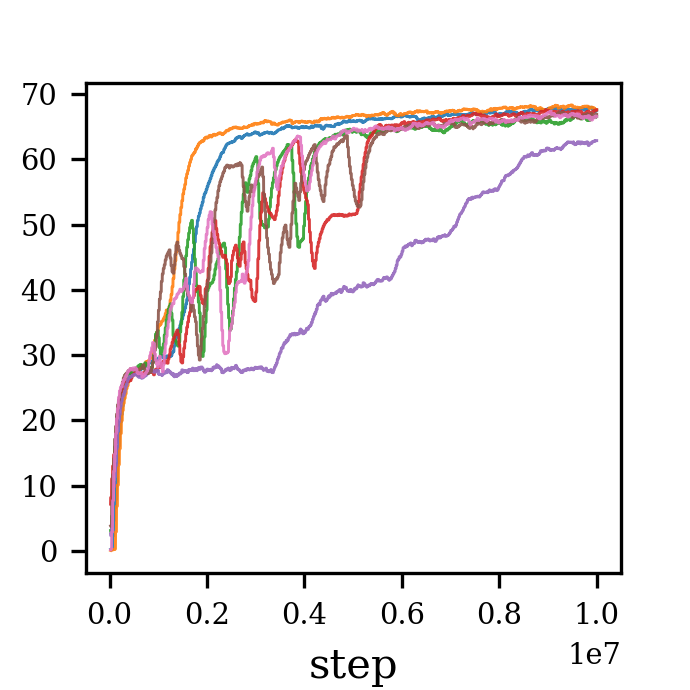}
     \caption{Freeway}
     \end{subfigure}
     \begin{subfigure}[b]{0.48\columnwidth}
     \includegraphics[width=\columnwidth]{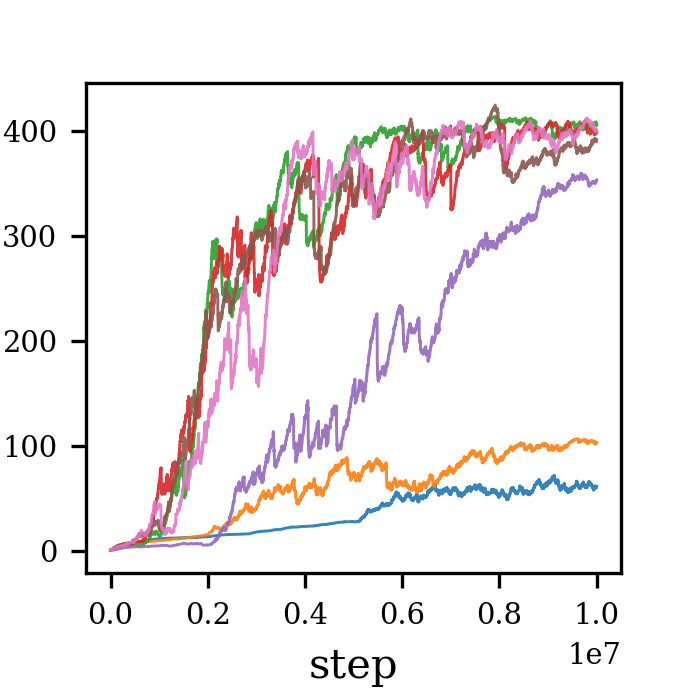}
     \caption{Breakout}
     \end{subfigure}
     \begin{subfigure}[b]{0.48\columnwidth}
     \includegraphics[width=\columnwidth]{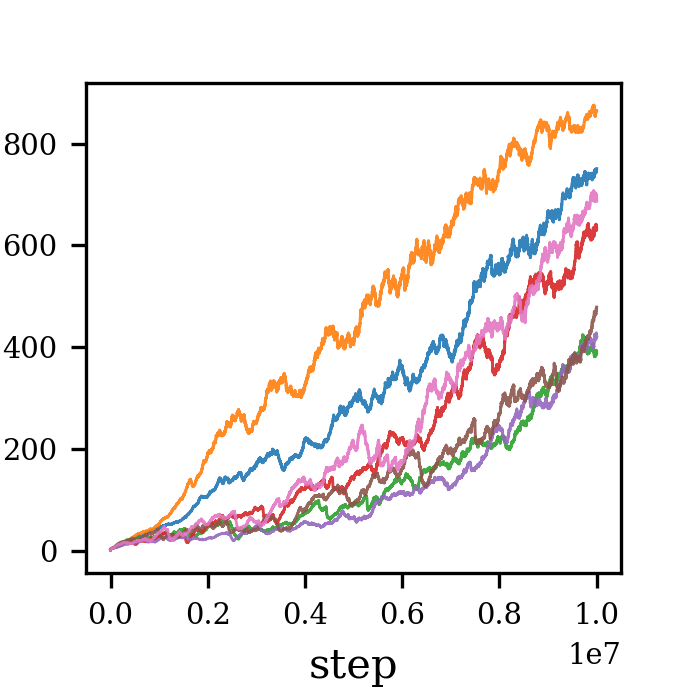}
     \caption{Space Invaders}
     \end{subfigure}
     \caption{Ablation of R-GCN-GTG on MinAtar tasks.}
     \label{fig:gcn_ablation}
     %\vspace{-10pt}
\end{figure}

Although the NLM-GTG models in our experiments make use of relational graphs determined by GTG, it uses this information in a less structured way. Specifically, NLM-GTGs encode such relational information as dense vector representations, whereas R-GCN-GTG uses this information to construct a GNN, thus directly determining the computation graph and flow of messages.
The performance gain of R-GCN-GTG over NLM-GTG suggests that using GTG to determine the specific relational inductive bias, and therefore guide message passing in a structured way, results in better in-distribution performance.
% This is sensible as the design of GTG  the logic of message passing constraints rather than merely the spatial information represented by a KB.

\subsection{Out-of-Distribution Systematic Generalization}
In \cref{tab:lava gen} and \cref{tab:rtfm}, we show how policies learned by our relational models can generalize to environments outside of the training distribution.
For our LavaCrossing experiments, we train the agent on difficulty level 2 and test the policy on difficulty levels 1 and 3.
\cref{tab:lava gen} shows average returns over five training runs. Each model is evaluated on 200 test episodes.
% During testing time, all the agent will choose the most likely actions rather than sample an action according to the output distribution, which improves the performance of R-GCN and CNN.
The relative performance change with respect to the training environment is shown in parentheses.
All the models generalize and perform optimally on difficulty level 1.
However, when generalizing to difficulty level 3, the relational models perform significantly better than CNNs.
R-GCN-GTG generalizes the best among all the models we tested.

\cref{tab:rtfm} demonstrates the win rate for each model in the symbolic variant of the RTFM tasks.
Again, we report mean returns averaged over five training runs.
Each section of the table represents a different task variation.
We transform the text document into a symbolic knowledge base of triples for RTFM-KB and RTFM-onehop-KB. This makes the task easier compared to the original RTFM-text task (last row) as models do not have to learn to encode information presented as textual inputs.
Further, the RTFM-onehop-KB is easier than RTFM-KB as the RTFM-KB require multihop reasoning.
We also put the performance of the model proposed by the RTFM paper~\citep{zhong2019rtfm}, $\text{txt}2\pi$ in RTFM-text environment into the table.
We train on environments with a grid size of 6$\times$6 and test generalization performance on environments with a grid size of 10$\times$10.
An optimal policy in the 10$\times$10 environments should achieve better performance compared to that on the smaller environments, as the agent has more space to evade monsters.
% However, we can see that the $\text{txt}2\pi$ only managed to keep the same level of performance when generalizing to larger fields.
We observe that NLM-GTG and R-GCN-GTG generalize well to the larger environments. However, R-GCN-GTG performs much better than NLM-GTG in the harder RTFM-KB environment, both in terms of in-distribution (6$\times$6 grid) and out-of-distribution generalization (10$\times$10 grid).

\begin{table}[t]
    \centering
    \small
\begin{tabular}{lcccc}
\toprule
 {\bf Model} &  {\bf Level 2} &  {\bf Level 1} &  {\bf Level 3} &  {\bf Portal} \\
\midrule
   CNN &   0.958 &   0.960 &   0.790(-17.5\%) &   0.040(-95.8\%) \\
   NLM-GTG &   0.955 &   0.960 &   0.918(-3.9\%) &   \textbf{0.158}(-83.5\%) \\
 R-GCN-GTG &   0.958 &   0.960 &   \textbf{0.942}(-1.7\%) &   0.096(-90.0\%)  \\
\bottomrule
\end{tabular}
    \caption{Out-of-distribution generalization performance on LavaCrossing. The agent is trained on difficulty level 2.}
    \label{tab:lava gen}
\end{table}
\subsection{Incorporating External knowledge}
The Portal-LavaCrossing and RTFM experiments demonstrate the flexibility of GTG in incorporating different kinds of external knowledge.
In \cref{tab:lava gen}, we can see that with spatial information provided by the KB, the NLM-GTG and R-GCN-GTG agent managed to generalize to Portal-LavaCrossing in a zero-shot manner.
The RTFM results in \cref{tab:rtfm} show how GTG enables the relational model to jointly reasoning with both spatial information and environment dynamics information when represented as a KB.
The R-GCN-GTG agent performs well both in multi-hop reasoning and one-hop reasoning variants of RTFM-KB, but NLM-GTG only performs well in the easier one-hop variant.
%In RTFM-text task, the $\text{txt}2\pi$ is trained with curriculum training.
%However, in this work, we can train the relational model directly on the target task and the performance is quite robust.

\begin{table}[t]
\centering
\small
\begin{tabular}{llcl}
\toprule
            & room size &      $6\times 6$ &     $10\times10$ \\
task & model &        &\\
\midrule
\multirow{2}{*}{RTFM-KB} & NLM-GTG &  21.0\% &  19.6\% (-6.7\%) \\
            & R-GCN-GTG &  86.3\% &  96.6\% (+12.0\%) \\
\cline{1-4}
\multirow{2}{*}{RTFM-onehop-KB} & NLM-GTG &  93.0\% &  99.4\% (+6.9\%) \\
            & R-GCN-GTG &  93.0\% &  98.2\% (+5.6\%) \\
\cline{1-4}
\multirow{1}{*}{RTFM-text} & $\text{txt}2\pi$ & 55\% & 55\% (+0\%) \\
\bottomrule
\end{tabular}
\caption{In-distribution and out-of-distribution generalization in RTFM variants. Figures report the win rate increment between $10\times10$ environments and $6\times6$ environments}
\label{tab:rtfm}
\end{table}

\subsection{Ablation Study}
\cref{fig:gcn_ablation} presents the results of an ablation study of three relational inductive biases encoded using GTG (local, remote, and auxiliary relations). Each line shows the smoothed training curve averaged over five training runs.
The red lines represent the training curve of R-GCN using the full set (local, remote, and auxiliary relations) of relational inductive biases.
The green lines show the performance of R-GCNs using local directional relations only, whose convolution computation is equivalent to that of the image convolution in \cref{sec:gtg}.
A notable difference among these models is the information aggregation method between convolution layers and dense layers: the CNN model concatenates all output feature vectors (i.e. flattened), while the R-GCN model applies a max-pooling layer.
The flattened vector in the CNN model tends to have high dimensionality, thereby increasing the total number of parameters in the adjacent dense layer.
Therefore, when constraining the number of parameters of the two architectures to be approximately equal, most of the parameters of the R-GCN model reside in convolution layers, whereas more parameters of the CNN model reside in the dense layers.
This explains the performance difference between CNN and R-GCN-GTG with local directional relations only.
In Seaquest and Breakout, R-GCNs yield better performance than CNNs, while in Space Invaders CNNs outperform R-GCN-GTG.
The two models achieve similar performance in Asterix and Freeway.

To further investigate the role of max-pooling, we evaluated a wider CNN with the same number of parameters as the convolution layer in the local-only R-GCN-GTG model. This wider model flattens features before dense layers rather than applying max-pooling, resulting in 876k parameters (the full R-GCN-GTG model has 149k).
We use this wider CNN model to isolate the performance improvement that results from max-pooling.
%There are two hypotheses about why max-pooling can help improve performance.
Comparing CNN-wide and R-GCN-GTG local-only, we see mixed results: the two models achieve comparable performance on Seaquest, Asterix and Freeway; local-only R-GCN-GTG performs better in Breakout, while the wider CNN model performs better in Space Invaders.
This shows that max-pooling by itself does not outperform flattening if we do not care about the number of parameters.

Unsurprisingly, removing local directional relations undermines the performance of R-GCN-GTG in almost all of the tasks, which shows the importance of the relational inductive bias of locality.
We also assessed the impact of remote and auxiliary relations, observing that using both sets of relations improves performance on Seaquest, Asterix and Space Invaders.
The benefits of introducing these additional relational inductive biases are robust in the sense that they do not degrade performance any environment. In contrast, this is not true for max-pooling.
These improvements demonstrate that we can go beyond the relational inductive bias of CNNs by using the relation determination rules of GTG, which provide a flexible framework for expressing many useful connectivity and parameter sharing constraints

\section{Conclusion}
This paper introduced Grid-to-Graph, a principled framework for representing relational inductive biases.
GTG is based on a set of relation determination rules, which act on inputs in the form of a feature map corresponding to discrete 2D state observations.
Using these relation determination rules, GTG transforms 2D observations into a multigraph input for an R-GCN model.
The resulting architecture, R-GCN-GTG outperforms both CNNs and Neural Logic Machines, the previous state-of-the-art in deep relational RL, on MinAtar and a series of challenging procedurally-generated grid world environments, both in terms of in-distribution performance and out-of-distribution systematic generalization. Our results further show that GTG provides an effective and straightforward interface for incorporating various forms of external knowledge without any architectural modifications.
%Our idea of encoding relational inductive biases as a relational graph is quite flexible, and GTG is one instantiation of this general idea for 2D feature-map inputs.
%One can either extend the relation determination rules of GTG, or designing novel relational inductive biases and GTG-like frameworks for other input types.
%Regarding GTG itself, a natural extension of it is to evaluate how R-GCN-GTG performs in settings with raw pixel inputs, such as playing Atari games and various computer vision tasks.

%
%\subsubsection*{Acknowledgments}
\begin{acks}
This research was supported by the European Union's Horizon 2020 research and innovation programme under grant agreement no. 875160.
We thank Edward Grefenstette and the anonymous reviewers for their insightful feedback. % to this paper.
\end{acks}

\bibliography{aamas}

%%% -*-BibTeX-*-
%%% Do NOT edit. File created by BibTeX with style
%%% ACM-Reference-Format-Journals [18-Jan-2012].

\begin{thebibliography}{37}

%%% ====================================================================
%%% NOTE TO THE USER: you can override these defaults by providing
%%% customized versions of any of these macros before the \bibliography
%%% command.  Each of them MUST provide its own final punctuation,
%%% except for \shownote{}, \showDOI{}, and \showURL{}.  The latter two
%%% do not use final punctuation, in order to avoid confusing it with
%%% the Web address.
%%%
%%% To suppress output of a particular field, define its macro to expand
%%% to an empty string, or better, \unskip, like this:
%%%
%%% \newcommand{\showDOI}[1]{\unskip}   % LaTeX syntax
%%%
%%% \def \showDOI #1{\unskip}           % plain TeX syntax
%%%
%%% ====================================================================

\ifx \showCODEN    \undefined \def \showCODEN     #1{\unskip}     \fi
\ifx \showDOI      \undefined \def \showDOI       #1{#1}\fi
\ifx \showISBNx    \undefined \def \showISBNx     #1{\unskip}     \fi
\ifx \showISBNxiii \undefined \def \showISBNxiii  #1{\unskip}     \fi
\ifx \showISSN     \undefined \def \showISSN      #1{\unskip}     \fi
\ifx \showLCCN     \undefined \def \showLCCN      #1{\unskip}     \fi
\ifx \shownote     \undefined \def \shownote      #1{#1}          \fi
\ifx \showarticletitle \undefined \def \showarticletitle #1{#1}   \fi
\ifx \showURL      \undefined \def \showURL       {\relax}        \fi
% The following commands are used for tagged output and should be
% invisible to TeX
\providecommand\bibfield[2]{#2}
\providecommand\bibinfo[2]{#2}
\providecommand\natexlab[1]{#1}
\providecommand\showeprint[2][]{arXiv:#2}

\bibitem[\protect\citeauthoryear{Battaglia, Hamrick, Bapst, Sanchez{-}Gonzalez,
  Zambaldi, Malinowski, Tacchetti, Raposo, Santoro, Faulkner,
  G{\"{u}}l{\c{c}}ehre, Song, Ballard, Gilmer, Dahl, Vaswani, Allen, Nash,
  Langston, Dyer, Heess, Wierstra, Kohli, Botvinick, Vinyals, Li, and
  Pascanu}{Battaglia et~al\mbox{.}}{2018}]%
        {battaglia2018relational}
\bibfield{author}{\bibinfo{person}{Peter~W. Battaglia},
  \bibinfo{person}{Jessica~B. Hamrick}, \bibinfo{person}{Victor Bapst},
  \bibinfo{person}{Alvaro Sanchez{-}Gonzalez},
  \bibinfo{person}{Vin{\'{\i}}cius~Flores Zambaldi}, \bibinfo{person}{Mateusz
  Malinowski}, \bibinfo{person}{Andrea Tacchetti}, \bibinfo{person}{David
  Raposo}, \bibinfo{person}{Adam Santoro}, \bibinfo{person}{Ryan Faulkner},
  \bibinfo{person}{{\c{C}}aglar G{\"{u}}l{\c{c}}ehre},
  \bibinfo{person}{H.~Francis Song}, \bibinfo{person}{Andrew~J. Ballard},
  \bibinfo{person}{Justin Gilmer}, \bibinfo{person}{George~E. Dahl},
  \bibinfo{person}{Ashish Vaswani}, \bibinfo{person}{Kelsey~R. Allen},
  \bibinfo{person}{Charles Nash}, \bibinfo{person}{Victoria Langston},
  \bibinfo{person}{Chris Dyer}, \bibinfo{person}{Nicolas Heess},
  \bibinfo{person}{Daan Wierstra}, \bibinfo{person}{Pushmeet Kohli},
  \bibinfo{person}{Matthew Botvinick}, \bibinfo{person}{Oriol Vinyals},
  \bibinfo{person}{Yujia Li}, {and} \bibinfo{person}{Razvan Pascanu}.}
  \bibinfo{year}{2018}\natexlab{}.
\newblock \showarticletitle{Relational inductive biases, deep learning, and
  graph networks}.
\newblock \bibinfo{journal}{\emph{CoRR}}  \bibinfo{volume}{abs/1806.01261}
  (\bibinfo{year}{2018}).
\newblock


\bibitem[\protect\citeauthoryear{Chevalier-Boisvert, Willems, and
  Pal}{Chevalier-Boisvert et~al\mbox{.}}{2018}]%
        {gym_minigrid}
\bibfield{author}{\bibinfo{person}{Maxime Chevalier-Boisvert},
  \bibinfo{person}{Lucas Willems}, {and} \bibinfo{person}{Suman Pal}.}
  \bibinfo{year}{2018}\natexlab{}.
\newblock \bibinfo{title}{Minimalistic Gridworld Environment for OpenAI Gym}.
\newblock
  \bibinfo{howpublished}{\url{https://github.com/maximecb/gym-minigrid}}.
\newblock


\bibitem[\protect\citeauthoryear{Cobbe, Hesse, Hilton, and Schulman}{Cobbe
  et~al\mbox{.}}{2020}]%
        {cobbe2020}
\bibfield{author}{\bibinfo{person}{Karl Cobbe}, \bibinfo{person}{Christopher
  Hesse}, \bibinfo{person}{Jacob Hilton}, {and} \bibinfo{person}{John
  Schulman}.} \bibinfo{year}{2020}\natexlab{}.
\newblock \showarticletitle{Leveraging Procedural Generation to Benchmark
  Reinforcement Learning}. In \bibinfo{booktitle}{\emph{Proceedings of the 37th
  International Conference on Machine Learning, {ICML} 2020, 13-18 July 2020,
  Virtual Event}} \emph{(\bibinfo{series}{Proceedings of Machine Learning
  Research}, Vol.~\bibinfo{volume}{119})}. \bibinfo{publisher}{{PMLR}},
  \bibinfo{pages}{2048--2056}.
\newblock
\urldef\tempurl%
\url{http://proceedings.mlr.press/v119/cobbe20a.html}
\showURL{%
\tempurl}


\bibitem[\protect\citeauthoryear{Cohen and Welling}{Cohen and Welling}{2016}]%
        {CohenW16}
\bibfield{author}{\bibinfo{person}{Taco Cohen} {and} \bibinfo{person}{Max
  Welling}.} \bibinfo{year}{2016}\natexlab{}.
\newblock \showarticletitle{Group Equivariant Convolutional Networks}. In
  \bibinfo{booktitle}{\emph{Proceedings of the 33nd International Conference on
  Machine Learning, {ICML} 2016, New York City, NY, USA, June 19-24, 2016}}
  \emph{(\bibinfo{series}{{JMLR} Workshop and Conference Proceedings},
  Vol.~\bibinfo{volume}{48})},
  \bibfield{editor}{\bibinfo{person}{Maria{-}Florina Balcan} {and}
  \bibinfo{person}{Kilian~Q. Weinberger}} (Eds.).
  \bibinfo{publisher}{JMLR.org}, \bibinfo{pages}{2990--2999}.
\newblock
\urldef\tempurl%
\url{http://proceedings.mlr.press/v48/cohenc16.html}
\showURL{%
\tempurl}


\bibitem[\protect\citeauthoryear{Degris, Pilarski, and Sutton}{Degris
  et~al\mbox{.}}{2012}]%
        {degris2012model}
\bibfield{author}{\bibinfo{person}{Thomas Degris}, \bibinfo{person}{Patrick~M
  Pilarski}, {and} \bibinfo{person}{Richard~S Sutton}.}
  \bibinfo{year}{2012}\natexlab{}.
\newblock \showarticletitle{Model-free reinforcement learning with continuous
  action in practice}. In \bibinfo{booktitle}{\emph{2012 American Control
  Conference (ACC)}}. IEEE, \bibinfo{pages}{2177--2182}.
\newblock


\bibitem[\protect\citeauthoryear{Dong, Mao, Lin, Wang, Li, and Zhou}{Dong
  et~al\mbox{.}}{2019}]%
        {dong2018neural}
\bibfield{author}{\bibinfo{person}{Honghua Dong}, \bibinfo{person}{Jiayuan
  Mao}, \bibinfo{person}{Tian Lin}, \bibinfo{person}{Chong Wang},
  \bibinfo{person}{Lihong Li}, {and} \bibinfo{person}{Denny Zhou}.}
  \bibinfo{year}{2019}\natexlab{}.
\newblock \showarticletitle{Neural Logic Machines}. In
  \bibinfo{booktitle}{\emph{{ICLR} (Poster)}}.
  \bibinfo{publisher}{OpenReview.net}.
\newblock


\bibitem[\protect\citeauthoryear{Espeholt, Soyer, Munos, Simonyan, Mnih, Ward,
  Doron, Firoiu, Harley, Dunning, Legg, and Kavukcuoglu}{Espeholt
  et~al\mbox{.}}{2018}]%
        {espeholt2018impala}
\bibfield{author}{\bibinfo{person}{Lasse Espeholt}, \bibinfo{person}{Hubert
  Soyer}, \bibinfo{person}{R{\'{e}}mi Munos}, \bibinfo{person}{Karen Simonyan},
  \bibinfo{person}{Volodymyr Mnih}, \bibinfo{person}{Tom Ward},
  \bibinfo{person}{Yotam Doron}, \bibinfo{person}{Vlad Firoiu},
  \bibinfo{person}{Tim Harley}, \bibinfo{person}{Iain Dunning},
  \bibinfo{person}{Shane Legg}, {and} \bibinfo{person}{Koray Kavukcuoglu}.}
  \bibinfo{year}{2018}\natexlab{}.
\newblock \showarticletitle{{IMPALA:} Scalable Distributed Deep-RL with
  Importance Weighted Actor-Learner Architectures}.
\newblock   \bibinfo{volume}{80} (\bibinfo{year}{2018}),
  \bibinfo{pages}{1406--1415}.
\newblock


\bibitem[\protect\citeauthoryear{Evans and Grefenstette}{Evans and
  Grefenstette}{2018}]%
        {Evans2018}
\bibfield{author}{\bibinfo{person}{Richard Evans} {and} \bibinfo{person}{Edward
  Grefenstette}.} \bibinfo{year}{2018}\natexlab{}.
\newblock \showarticletitle{{Learning Explanatory Rules from Noisy Data}}.
\newblock \bibinfo{journal}{\emph{Journal of Artificial Intelligence Research}}
   \bibinfo{volume}{61} (\bibinfo{date}{Jan} \bibinfo{year}{2018}),
  \bibinfo{pages}{1--64}.
\newblock
\showISSN{1076-9757}
\urldef\tempurl%
\url{https://doi.org/10.1613/jair.5714}
\showDOI{\tempurl}


\bibitem[\protect\citeauthoryear{Farebrother, Machado, and Bowling}{Farebrother
  et~al\mbox{.}}{2018}]%
        {Farebrother2018}
\bibfield{author}{\bibinfo{person}{Jesse Farebrother},
  \bibinfo{person}{Marlos~C. Machado}, {and} \bibinfo{person}{Michael
  Bowling}.} \bibinfo{year}{2018}\natexlab{}.
\newblock \showarticletitle{Generalization and Regularization in {DQN}}.
\newblock \bibinfo{journal}{\emph{CoRR}}  \bibinfo{volume}{abs/1810.00123}
  (\bibinfo{year}{2018}).
\newblock
\showeprint[arxiv]{1810.00123}
\urldef\tempurl%
\url{http://arxiv.org/abs/1810.00123}
\showURL{%
\tempurl}


\bibitem[\protect\citeauthoryear{Fey and Lenssen}{Fey and Lenssen}{2019}]%
        {Fey/Lenssen/2019}
\bibfield{author}{\bibinfo{person}{Matthias Fey} {and} \bibinfo{person}{Jan~E.
  Lenssen}.} \bibinfo{year}{2019}\natexlab{}.
\newblock \showarticletitle{Fast Graph Representation Learning with {PyTorch
  Geometric}}. In \bibinfo{booktitle}{\emph{ICLR Workshop on Representation
  Learning on Graphs and Manifolds}}.
\newblock


\bibitem[\protect\citeauthoryear{Goodfellow, Bengio, Courville, and
  Bengio}{Goodfellow et~al\mbox{.}}{2016}]%
        {goodfellow2016deep}
\bibfield{author}{\bibinfo{person}{Ian Goodfellow}, \bibinfo{person}{Yoshua
  Bengio}, \bibinfo{person}{Aaron Courville}, {and} \bibinfo{person}{Yoshua
  Bengio}.} \bibinfo{year}{2016}\natexlab{}.
\newblock \bibinfo{booktitle}{\emph{Deep learning}}. Vol.~\bibinfo{volume}{1}.
\newblock \bibinfo{publisher}{MIT press Cambridge}.
\newblock


\bibitem[\protect\citeauthoryear{Hausknecht and Stone}{Hausknecht and
  Stone}{2015}]%
        {hausknecht2015impact}
\bibfield{author}{\bibinfo{person}{Matthew~J Hausknecht} {and}
  \bibinfo{person}{Peter Stone}.} \bibinfo{year}{2015}\natexlab{}.
\newblock \showarticletitle{The Impact of Determinism on Learning Atari 2600
  Games.}. In \bibinfo{booktitle}{\emph{AAAI Workshop: Learning for General
  Competency in Video Games}}.
\newblock


\bibitem[\protect\citeauthoryear{Hochreiter and Schmidhuber}{Hochreiter and
  Schmidhuber}{1997}]%
        {hochreiter1997long}
\bibfield{author}{\bibinfo{person}{Sepp Hochreiter} {and}
  \bibinfo{person}{J{\"u}rgen Schmidhuber}.} \bibinfo{year}{1997}\natexlab{}.
\newblock \showarticletitle{Long short-term memory}.
\newblock \bibinfo{journal}{\emph{Neural computation}} \bibinfo{volume}{9},
  \bibinfo{number}{8} (\bibinfo{year}{1997}), \bibinfo{pages}{1735--1780}.
\newblock


\bibitem[\protect\citeauthoryear{Jiang and Luo}{Jiang and Luo}{2019}]%
        {pmlr-v97-jiang19a}
\bibfield{author}{\bibinfo{person}{Zhengyao Jiang} {and} \bibinfo{person}{Shan
  Luo}.} \bibinfo{year}{2019}\natexlab{}.
\newblock \showarticletitle{Neural Logic Reinforcement Learning}. In
  \bibinfo{booktitle}{\emph{Proceedings of the 36th International Conference on
  Machine Learning}} \emph{(\bibinfo{series}{Proceedings of Machine Learning
  Research}, Vol.~\bibinfo{volume}{97})},
  \bibfield{editor}{\bibinfo{person}{Kamalika Chaudhuri} {and}
  \bibinfo{person}{Ruslan Salakhutdinov}} (Eds.). \bibinfo{publisher}{PMLR},
  \bibinfo{address}{Long Beach, California, USA}, \bibinfo{pages}{3110--3119}.
\newblock
\urldef\tempurl%
\url{http://proceedings.mlr.press/v97/jiang19a.html}
\showURL{%
\tempurl}


\bibitem[\protect\citeauthoryear{Krizhevsky, Sutskever, and Hinton}{Krizhevsky
  et~al\mbox{.}}{2012a}]%
        {krizhevsky2012imagenet}
\bibfield{author}{\bibinfo{person}{Alex Krizhevsky}, \bibinfo{person}{Ilya
  Sutskever}, {and} \bibinfo{person}{Geoffrey~E Hinton}.}
  \bibinfo{year}{2012}\natexlab{a}.
\newblock \showarticletitle{Imagenet classification with deep convolutional
  neural networks}. In \bibinfo{booktitle}{\emph{Advances in neural information
  processing systems}}. \bibinfo{pages}{1097--1105}.
\newblock


\bibitem[\protect\citeauthoryear{Krizhevsky, Sutskever, and Hinton}{Krizhevsky
  et~al\mbox{.}}{2012b}]%
        {NIPS2012_4824}
\bibfield{author}{\bibinfo{person}{Alex Krizhevsky}, \bibinfo{person}{Ilya
  Sutskever}, {and} \bibinfo{person}{Geoffrey~E Hinton}.}
  \bibinfo{year}{2012}\natexlab{b}.
\newblock \showarticletitle{ImageNet Classification with Deep Convolutional
  Neural Networks}.
\newblock In \bibinfo{booktitle}{\emph{Advances in Neural Information
  Processing Systems 25}}, \bibfield{editor}{\bibinfo{person}{F.~Pereira},
  \bibinfo{person}{C.~J.~C. Burges}, \bibinfo{person}{L.~Bottou}, {and}
  \bibinfo{person}{K.~Q. Weinberger}} (Eds.). \bibinfo{publisher}{Curran
  Associates, Inc.}, \bibinfo{pages}{1097--1105}.
\newblock
\urldef\tempurl%
\url{http://papers.nips.cc/paper/4824-imagenet-classification-with-deep-convolutional-neural-networks.pdf}
\showURL{%
\tempurl}


\bibitem[\protect\citeauthoryear{Kurin, Igl, Rockt{\"a}schel, Boehmer, and
  Whiteson}{Kurin et~al\mbox{.}}{2020}]%
        {kurin2020body}
\bibfield{author}{\bibinfo{person}{Vitaly Kurin}, \bibinfo{person}{Maximilian
  Igl}, \bibinfo{person}{Tim Rockt{\"a}schel}, \bibinfo{person}{Wendelin
  Boehmer}, {and} \bibinfo{person}{Shimon Whiteson}.}
  \bibinfo{year}{2020}\natexlab{}.
\newblock \showarticletitle{My Body is a Cage: the Role of Morphology in
  Graph-Based Incompatible Control}. In \bibinfo{booktitle}{\emph{International
  Conference on Learaning Representations (ICLR)}}.
\newblock


\bibitem[\protect\citeauthoryear{K\"{u}ttler, Nardelli, Lavril, Selvatici,
  Sivakumar, Rockt\"{a}schel, and Grefenstette}{K\"{u}ttler
  et~al\mbox{.}}{2019}]%
        {torchbeast2019}
\bibfield{author}{\bibinfo{person}{Heinrich K\"{u}ttler},
  \bibinfo{person}{Nantas Nardelli}, \bibinfo{person}{Thibaut Lavril},
  \bibinfo{person}{Marco Selvatici}, \bibinfo{person}{Viswanath Sivakumar},
  \bibinfo{person}{Tim Rockt\"{a}schel}, {and} \bibinfo{person}{Edward
  Grefenstette}.} \bibinfo{year}{2019}\natexlab{}.
\newblock \showarticletitle{{TorchBeast: A PyTorch Platform for Distributed
  RL}}.
\newblock \bibinfo{journal}{\emph{arXiv preprint arXiv:1910.03552}}
  (\bibinfo{year}{2019}).
\newblock
\urldef\tempurl%
\url{https://github.com/facebookresearch/torchbeast}
\showURL{%
\tempurl}


\bibitem[\protect\citeauthoryear{Mnih, Kavukcuoglu, Silver, Rusu, Veness,
  Bellemare, Graves, Riedmiller, Fidjeland, Ostrovski, et~al\mbox{.}}{Mnih
  et~al\mbox{.}}{2015a}]%
        {mnih2015human}
\bibfield{author}{\bibinfo{person}{Volodymyr Mnih}, \bibinfo{person}{Koray
  Kavukcuoglu}, \bibinfo{person}{David Silver}, \bibinfo{person}{Andrei~A
  Rusu}, \bibinfo{person}{Joel Veness}, \bibinfo{person}{Marc~G Bellemare},
  \bibinfo{person}{Alex Graves}, \bibinfo{person}{Martin Riedmiller},
  \bibinfo{person}{Andreas~K Fidjeland}, \bibinfo{person}{Georg Ostrovski},
  {et~al\mbox{.}}} \bibinfo{year}{2015}\natexlab{a}.
\newblock \showarticletitle{Human-level control through deep reinforcement
  learning}.
\newblock \bibinfo{journal}{\emph{nature}} \bibinfo{volume}{518},
  \bibinfo{number}{7540} (\bibinfo{year}{2015}), \bibinfo{pages}{529--533}.
\newblock


\bibitem[\protect\citeauthoryear{Mnih, Kavukcuoglu, Silver, Rusu, Veness,
  Bellemare, Graves, Riedmiller, Fidjeland, Ostrovski, Petersen, Beattie,
  Sadik, Antonoglou, King, Kumaran, Wierstra, Legg, and Hassabis}{Mnih
  et~al\mbox{.}}{2015b}]%
        {Mnih2015}
\bibfield{author}{\bibinfo{person}{Volodymyr Mnih}, \bibinfo{person}{Koray
  Kavukcuoglu}, \bibinfo{person}{David Silver}, \bibinfo{person}{Andrei~A.
  Rusu}, \bibinfo{person}{Joel Veness}, \bibinfo{person}{Marc~G. Bellemare},
  \bibinfo{person}{Alex Graves}, \bibinfo{person}{Martin Riedmiller},
  \bibinfo{person}{Andreas~K. Fidjeland}, \bibinfo{person}{Georg Ostrovski},
  \bibinfo{person}{Stig Petersen}, \bibinfo{person}{Charles Beattie},
  \bibinfo{person}{Amir Sadik}, \bibinfo{person}{Ioannis Antonoglou},
  \bibinfo{person}{Helen King}, \bibinfo{person}{Dharshan Kumaran},
  \bibinfo{person}{Daan Wierstra}, \bibinfo{person}{Shane Legg}, {and}
  \bibinfo{person}{Demis Hassabis}.} \bibinfo{year}{2015}\natexlab{b}.
\newblock \showarticletitle{{Human-level control through deep reinforcement
  learning}}.
\newblock \bibinfo{journal}{\emph{Nature}} \bibinfo{volume}{518},
  \bibinfo{number}{7540} (\bibinfo{date}{Feb} \bibinfo{year}{2015}),
  \bibinfo{pages}{529--533}.
\newblock
\showISBNx{1476-4687 (Electronic) 0028-0836 (Linking)}
\showISSN{0028-0836}
\urldef\tempurl%
\url{https://doi.org/10.1038/nature14236}
\showDOI{\tempurl}
\showeprint{1312.5602}


\bibitem[\protect\citeauthoryear{P{\"{u}}schel and Moura}{P{\"{u}}schel and
  Moura}{2008}]%
        {PuschelM08a}
\bibfield{author}{\bibinfo{person}{Markus P{\"{u}}schel} {and}
  \bibinfo{person}{Jos{\'{e}} M.~F. Moura}.} \bibinfo{year}{2008}\natexlab{}.
\newblock \showarticletitle{Algebraic Signal Processing Theory: Foundation and
  1-D Time}.
\newblock \bibinfo{journal}{\emph{{IEEE} Trans. Signal Process.}}
  \bibinfo{volume}{56}, \bibinfo{number}{8-1} (\bibinfo{year}{2008}),
  \bibinfo{pages}{3572--3585}.
\newblock
\urldef\tempurl%
\url{https://doi.org/10.1109/TSP.2008.925261}
\showDOI{\tempurl}


\bibitem[\protect\citeauthoryear{Schlichtkrull, Kipf, Bloem, Van Den~Berg,
  Titov, and Welling}{Schlichtkrull et~al\mbox{.}}{2018}]%
        {schlichtkrull2018modeling}
\bibfield{author}{\bibinfo{person}{Michael Schlichtkrull},
  \bibinfo{person}{Thomas~N Kipf}, \bibinfo{person}{Peter Bloem},
  \bibinfo{person}{Rianne Van Den~Berg}, \bibinfo{person}{Ivan Titov}, {and}
  \bibinfo{person}{Max Welling}.} \bibinfo{year}{2018}\natexlab{}.
\newblock \showarticletitle{Modeling relational data with graph convolutional
  networks}. In \bibinfo{booktitle}{\emph{European Semantic Web Conference}}.
  Springer, \bibinfo{pages}{593--607}.
\newblock


\bibitem[\protect\citeauthoryear{Silver, Hubert, Schrittwieser, Antonoglou,
  Lai, Guez, Lanctot, Sifre, Kumaran, Graepel, et~al\mbox{.}}{Silver
  et~al\mbox{.}}{2018}]%
        {silver2018general}
\bibfield{author}{\bibinfo{person}{David Silver}, \bibinfo{person}{Thomas
  Hubert}, \bibinfo{person}{Julian Schrittwieser}, \bibinfo{person}{Ioannis
  Antonoglou}, \bibinfo{person}{Matthew Lai}, \bibinfo{person}{Arthur Guez},
  \bibinfo{person}{Marc Lanctot}, \bibinfo{person}{Laurent Sifre},
  \bibinfo{person}{Dharshan Kumaran}, \bibinfo{person}{Thore Graepel},
  {et~al\mbox{.}}} \bibinfo{year}{2018}\natexlab{}.
\newblock \showarticletitle{A general reinforcement learning algorithm that
  masters chess, shogi, and Go through self-play}.
\newblock \bibinfo{journal}{\emph{Science}} \bibinfo{volume}{362},
  \bibinfo{number}{6419} (\bibinfo{year}{2018}), \bibinfo{pages}{1140--1144}.
\newblock


\bibitem[\protect\citeauthoryear{Sukhbaatar, Szlam, Weston, and
  Fergus}{Sukhbaatar et~al\mbox{.}}{2015}]%
        {sukhbaatar2015end}
\bibfield{author}{\bibinfo{person}{Sainbayar Sukhbaatar},
  \bibinfo{person}{Arthur Szlam}, \bibinfo{person}{Jason Weston}, {and}
  \bibinfo{person}{Rob Fergus}.} \bibinfo{year}{2015}\natexlab{}.
\newblock \showarticletitle{End-To-End Memory Networks}. In
  \bibinfo{booktitle}{\emph{{NIPS}}}. \bibinfo{pages}{2440--2448}.
\newblock


\bibitem[\protect\citeauthoryear{Sutton and Barto}{Sutton and Barto}{1998}]%
        {SuttonBook}
\bibfield{author}{\bibinfo{person}{Richard~S. Sutton} {and}
  \bibinfo{person}{Andrew~G. Barto}.} \bibinfo{year}{1998}\natexlab{}.
\newblock \bibinfo{booktitle}{\emph{Introduction to Reinforcement Learning}
  (\bibinfo{edition}{1st} ed.)}.
\newblock \bibinfo{publisher}{MIT Press}, \bibinfo{address}{Cambridge, MA,
  USA}.
\newblock
\showISBNx{0262193981}


\bibitem[\protect\citeauthoryear{Tieleman and Hinton}{Tieleman and
  Hinton}{2012}]%
        {Tieleman2012}
\bibfield{author}{\bibinfo{person}{T. Tieleman} {and} \bibinfo{person}{G.
  Hinton}.} \bibinfo{year}{2012}\natexlab{}.
\newblock \bibinfo{title}{{Lecture 6.5---RmsProp: Divide the gradient by a
  running average of its recent magnitude}}.
\newblock \bibinfo{howpublished}{COURSERA: Neural Networks for Machine
  Learning}.
\newblock


\bibitem[\protect\citeauthoryear{van~der Pol, Worrall, van Hoof, Oliehoek, and
  Welling}{van~der Pol et~al\mbox{.}}{2020}]%
        {PolWHOW20}
\bibfield{author}{\bibinfo{person}{Elise van~der Pol},
  \bibinfo{person}{Daniel~E. Worrall}, \bibinfo{person}{Herke van Hoof},
  \bibinfo{person}{Frans~A. Oliehoek}, {and} \bibinfo{person}{Max Welling}.}
  \bibinfo{year}{2020}\natexlab{}.
\newblock \showarticletitle{{MDP} Homomorphic Networks: Group Symmetries in
  Reinforcement Learning}. In \bibinfo{booktitle}{\emph{Advances in Neural
  Information Processing Systems 33: Annual Conference on Neural Information
  Processing Systems 2020, NeurIPS 2020, December 6-12, 2020, virtual}},
  \bibfield{editor}{\bibinfo{person}{Hugo Larochelle},
  \bibinfo{person}{Marc'Aurelio Ranzato}, \bibinfo{person}{Raia Hadsell},
  \bibinfo{person}{Maria{-}Florina Balcan}, {and} \bibinfo{person}{Hsuan{-}Tien
  Lin}} (Eds.).
\newblock
\urldef\tempurl%
\url{https://proceedings.neurips.cc/paper/2020/hash/2be5f9c2e3620eb73c2972d7552b6cb5-Abstract.html}
\showURL{%
\tempurl}


\bibitem[\protect\citeauthoryear{Wang, Liao, Ba, and Fidler}{Wang
  et~al\mbox{.}}{2018}]%
        {Wang2018}
\bibfield{author}{\bibinfo{person}{Tingwu Wang}, \bibinfo{person}{Renjie Liao},
  \bibinfo{person}{Jimmy Ba}, {and} \bibinfo{person}{Sanja Fidler}.}
  \bibinfo{year}{2018}\natexlab{}.
\newblock \showarticletitle{NerveNet: Learning Structured Policy with Graph
  Neural Networks}. In \bibinfo{booktitle}{\emph{{ICLR} (Poster)}}.
  \bibinfo{publisher}{OpenReview.net}.
\newblock


\bibitem[\protect\citeauthoryear{{Young} and {Tian}}{{Young} and
  {Tian}}{2019}]%
        {Young2019}
\bibfield{author}{\bibinfo{person}{Kenny {Young}} {and} \bibinfo{person}{Tian
  {Tian}}.} \bibinfo{year}{2019}\natexlab{}.
\newblock \showarticletitle{MinAtar: An Atari-inspired Testbed for More
  Efficient Reinforcement Learning Experiments}.
\newblock \bibinfo{journal}{\emph{arXiv preprint arXiv:1903.03176}}
  (\bibinfo{year}{2019}).
\newblock


\bibitem[\protect\citeauthoryear{Young and Tian}{Young and Tian}{2019}]%
        {young2019minatar}
\bibfield{author}{\bibinfo{person}{Kenny Young} {and} \bibinfo{person}{Tian
  Tian}.} \bibinfo{year}{2019}\natexlab{}.
\newblock \showarticletitle{Minatar: An atari-inspired testbed for thorough and
  reproducible reinforcement learning experiments}.
\newblock \bibinfo{journal}{\emph{arXiv preprint arXiv:1903.03176}}
  (\bibinfo{year}{2019}).
\newblock


\bibitem[\protect\citeauthoryear{Yu and Koltun}{Yu and Koltun}{2016}]%
        {yu2015multi}
\bibfield{author}{\bibinfo{person}{Fisher Yu} {and} \bibinfo{person}{Vladlen
  Koltun}.} \bibinfo{year}{2016}\natexlab{}.
\newblock \showarticletitle{Multi-Scale Context Aggregation by Dilated
  Convolutions}.
\newblock  (\bibinfo{year}{2016}).
\newblock


\bibitem[\protect\citeauthoryear{Zaheer, Kottur, Ravanbakhsh, P{\'{o}}czos,
  Salakhutdinov, and Smola}{Zaheer et~al\mbox{.}}{2017}]%
        {zaheer2017deep}
\bibfield{author}{\bibinfo{person}{Manzil Zaheer}, \bibinfo{person}{Satwik
  Kottur}, \bibinfo{person}{Siamak Ravanbakhsh},
  \bibinfo{person}{Barnab{\'{a}}s P{\'{o}}czos}, \bibinfo{person}{Ruslan
  Salakhutdinov}, {and} \bibinfo{person}{Alexander~J. Smola}.}
  \bibinfo{year}{2017}\natexlab{}.
\newblock \showarticletitle{Deep Sets}. In \bibinfo{booktitle}{\emph{{NIPS}}}.
  \bibinfo{pages}{3391--3401}.
\newblock


\bibitem[\protect\citeauthoryear{{Zambaldi}, {Raposo}, {Santoro}, {Bapst},
  {Li}, {Babuschkin}, {Tuyls}, {Reichert}, {Lillicrap}, {Lockhart}, {Shanahan},
  {Langston}, {Pascanu}, {Botvinick}, {Vinyals}, and {Battaglia}}{{Zambaldi}
  et~al\mbox{.}}{2018}]%
        {zambaldi2018}
\bibfield{author}{\bibinfo{person}{Vinicius {Zambaldi}}, \bibinfo{person}{David
  {Raposo}}, \bibinfo{person}{Adam {Santoro}}, \bibinfo{person}{Victor
  {Bapst}}, \bibinfo{person}{Yujia {Li}}, \bibinfo{person}{Igor {Babuschkin}},
  \bibinfo{person}{Karl {Tuyls}}, \bibinfo{person}{David {Reichert}},
  \bibinfo{person}{Timothy {Lillicrap}}, \bibinfo{person}{Edward {Lockhart}},
  \bibinfo{person}{Murray {Shanahan}}, \bibinfo{person}{Victoria {Langston}},
  \bibinfo{person}{Razvan {Pascanu}}, \bibinfo{person}{Matthew {Botvinick}},
  \bibinfo{person}{Oriol {Vinyals}}, {and} \bibinfo{person}{Peter
  {Battaglia}}.} \bibinfo{year}{2018}\natexlab{}.
\newblock \showarticletitle{{Relational Deep Reinforcement Learning}}.
\newblock \bibinfo{journal}{\emph{arXiv preprint}}
  \bibinfo{volume}{abs/1806.01830} (\bibinfo{date}{June} \bibinfo{year}{2018}).
\newblock


\bibitem[\protect\citeauthoryear{Zambaldi, Raposo, Santoro, Bapst, Li,
  Babuschkin, Tuyls, Reichert, Lillicrap, Lockhart, Shanahan, Langston,
  Pascanu, Botvinick, Vinyals, and Battaglia}{Zambaldi et~al\mbox{.}}{2019}]%
        {zambaldi2018deep}
\bibfield{author}{\bibinfo{person}{Vin{\'{\i}}cius~Flores Zambaldi},
  \bibinfo{person}{David Raposo}, \bibinfo{person}{Adam Santoro},
  \bibinfo{person}{Victor Bapst}, \bibinfo{person}{Yujia Li},
  \bibinfo{person}{Igor Babuschkin}, \bibinfo{person}{Karl Tuyls},
  \bibinfo{person}{David~P. Reichert}, \bibinfo{person}{Timothy~P. Lillicrap},
  \bibinfo{person}{Edward Lockhart}, \bibinfo{person}{Murray Shanahan},
  \bibinfo{person}{Victoria Langston}, \bibinfo{person}{Razvan Pascanu},
  \bibinfo{person}{Matthew Botvinick}, \bibinfo{person}{Oriol Vinyals}, {and}
  \bibinfo{person}{Peter~W. Battaglia}.} \bibinfo{year}{2019}\natexlab{}.
\newblock \showarticletitle{Deep reinforcement learning with relational
  inductive biases}. In \bibinfo{booktitle}{\emph{{ICLR} (Poster)}}.
  \bibinfo{publisher}{OpenReview.net}.
\newblock


\bibitem[\protect\citeauthoryear{Zhang, Vinyals, Munos, and Bengio}{Zhang
  et~al\mbox{.}}{2018}]%
        {Zhang2018}
\bibfield{author}{\bibinfo{person}{Chiyuan Zhang}, \bibinfo{person}{Oriol
  Vinyals}, \bibinfo{person}{R{\'{e}}mi Munos}, {and} \bibinfo{person}{Samy
  Bengio}.} \bibinfo{year}{2018}\natexlab{}.
\newblock \showarticletitle{A Study on Overfitting in Deep Reinforcement
  Learning}.
\newblock \bibinfo{journal}{\emph{CoRR}}  \bibinfo{volume}{abs/1804.06893}
  (\bibinfo{year}{2018}).
\newblock
\showeprint[arxiv]{1804.06893}
\urldef\tempurl%
\url{http://arxiv.org/abs/1804.06893}
\showURL{%
\tempurl}


\bibitem[\protect\citeauthoryear{Zhong, Rockt{\"{a}}schel, and
  Grefenstette}{Zhong et~al\mbox{.}}{2020}]%
        {zhong2019rtfm}
\bibfield{author}{\bibinfo{person}{Victor Zhong}, \bibinfo{person}{Tim
  Rockt{\"{a}}schel}, {and} \bibinfo{person}{Edward Grefenstette}.}
  \bibinfo{year}{2020}\natexlab{}.
\newblock \showarticletitle{{RTFM:} Generalising to New Environment Dynamics
  via Reading}.
\newblock  (\bibinfo{year}{2020}).
\newblock


\bibitem[\protect\citeauthoryear{Zhou, Cui, Zhang, Yang, Liu, Wang, Li, and
  Sun}{Zhou et~al\mbox{.}}{2018}]%
        {zhou2018graph}
\bibfield{author}{\bibinfo{person}{Jie Zhou}, \bibinfo{person}{Ganqu Cui},
  \bibinfo{person}{Zhengyan Zhang}, \bibinfo{person}{Cheng Yang},
  \bibinfo{person}{Zhiyuan Liu}, \bibinfo{person}{Lifeng Wang},
  \bibinfo{person}{Changcheng Li}, {and} \bibinfo{person}{Maosong Sun}.}
  \bibinfo{year}{2018}\natexlab{}.
\newblock \showarticletitle{Graph neural networks: A review of methods and
  applications}.
\newblock \bibinfo{journal}{\emph{arXiv preprint arXiv:1812.08434}}
  (\bibinfo{year}{2018}).
\newblock


\end{thebibliography}
\bibliographystyle{ACM-Reference-Format}

\newpage

\phantomsection
\appendix
\section{Appendix}
\subsection{Hyperparameters}
All models used in our experiments, including relational models and the baseline CNN, have 4 hidden layers. 
We train all models using RMSprop~\citep{Tieleman2012} with a learning rate of 0.001.

In the RL setting, an agent can, in principle, produce an infinite amount of training data given enough computational power and time, implying that increasing the model size can almost always be helpful for in-distribution performance.
For fair comparison, we aimed to ensure similar parameter counts for all relational models tested.
Taking LavaCrossing as an example, the CNN model contains 149k total parameters; R-GCN, 149k; and NLM, 139k.

We also evaluated even larger CNN models but did not observe significant improvements to in-distribution performance.
We only tested NLM with maximum arity of 2, because of the large  computational cost of higher arity.
Each layer and each arity had an output dimension of 64, resulting in 192 intensional predicates per layer.
R-GCN models used 2 relational convolution layers, each with an outputting 64-dimensional feature vectors.
After max-pooling, we then apply 2 dense layers, each with 128 hidden units.
CNN models used 2 convolution layers, each with a $3\times3$ kernel and 12 features, and 2 dense layers, each with 128 hidden units.

\subsection{Relational Inductive Bias and Generalization}
Prior work ~\cite{battaglia2018relational} defines relational inductive bias as constraints on relationships and interactions among entities in a learning process.
Usually, the relational inductive bias is implemented by a specific pattern of parameter sharing and neural connectivity in the neural architecture.
For example, CNNs exploit local connections constraining information processing to a limited, local receptive field, sharing parameters between different local kernels to introduce spatial translation invariance.
Such relational inductive biases have been argued to be critical for promoting combinatorial generalization \cite{battaglia2018relational}.

Combinatorial generalization is the process of exploiting compositional structure underlying a problem to successfully perform inference, prediction, or useful behaviours on previously unseen examples or scenarios \cite{battaglia2018relational}.
This concept is closely related to systematic generalization, a defining capability human intelligence, which modern deep learning methods have yet to attain.
Combinatorial generalization can help the model improve sample efficiency and generalize to new tasks.
In practice, it is important to disentangle combinatorial generalization from memorization.
One way to explicitly test for combinatorial generalization is to test the model on out-of-distribution held-out data (sampled from a different distribution than the training data). This data should be generated by a process mirroring the compositional rules governing the generation of the training data.

Unlike in-distribution generalization, which has been well-studied by learning theory and statistics, out-of-distribution generalization cannot, in general, be achieved by simply increasing the amount of data. 
However, successful combinatorial generalization would allow out-of-distribution generalization on such held-out data that follows similar composition rules as the training data.

In the RL setting, given a fixed policy $\pi$, the sampled data consists of trajectories of states, actions and rewards.
We call an RL environment \emph{out-of-distribution} if, at test time, either initial state, dynamics, or the task distribution in multi-task learning, is varied such that the distribution of trajectories differs from that at training.
We say an environment or collection of environments (with some appropriate sampling distribution over these environments) is \emph{in-distribution} if it generates trajectories with the same likelihood as under the training distribution.
For example, consider the case in which the initial state of an environment is procedurally generated and, during testing, we keep all the level generation logic the same, only using different random seeds.
Though the agent may see initial states it never saw in training, we will would not call these test-time configurations out-of-distribution, as the probability that the agent meets these initial states remains the same at test time as at training time. 

\subsection{Matrix Representation of Message Passing} \label{sec:block}
Here we provide a block matrix formalization of a class of neural network layers, shedding light how different neural architectures can be represented by their respective relational inductive biases.
Suppose we have $n$ $m$-dimensional input feature vectors, $\mathbf{x}_1, \ldots, \mathbf{x}_n \in \mathbb{R}^m$, and want to map them to $n$ output feature vectors $\mathbf{y}_1, \ldots, \mathbf{y}_n \in \mathbb{R}^m$ of the same dimensionality.
If we consider the case where this mapping is linear, we can concatenate all the input feature vectors $\mathbf{x}_i$ and express the transformation as a block matrix product:
\begin{equation*}
\begin{aligned}
    \begin{bmatrix} 
    \mb{A}_{11} & \dots  &  \mb{A}_{1b} & \dots & \mb{A}_{1n}\\
    \vdots & \ddots &         &       & \vdots\\
    \mb{A}_{a1} &        & \mb{A}_{ab}  &       & \mb{A}_{an}\\
    \vdots &        &         & \ddots& \vdots\\
    \mb{A}_{n1} &  \dots & \mb{A}_{n}  & \dots & \mb{A}_{nn}\\
    \end{bmatrix}
    \begin{bmatrix}     
    \mb{x}_{1}  \\ 
    \vdots  \\
    \mb{x}_{b}  \\
    \vdots  \\
    \mb{x}_{n}  \\
    \end{bmatrix}
    =
    \begin{bmatrix}     
    \mb{y}_{1}  \\ 
    \vdots  \\
    \mb{y}_{a}  \\
    \vdots  \\
    \mb{y}_{n}  \\
    \end{bmatrix}
    \label{eq:block}
\end{aligned}
\end{equation*}
We see the update rule of $\mb{y}_a$ is:
\begin{equation} \label{eq:update}
\mb{y}_a=\sum_{b=1}^n \mb{A}_{ab}\mb{x}_b
\end{equation}
The submatrix $\mb{A}_{ab} \in \mathbb{R}^{m \times m}$, which we refer to as the \emph{message passing matrix}, dictates the message passing from  entity $a$ to entity $b$.
When no constraints are applied to $\mb{A}$, the overall mapping expressed in \cref{eq:block} corresponds to a standard dense linear layer.
In the following sections, we refer to the constraints over the message passing matrices $\mb{A}$ as the relational inductive bias.
This formalization provides a more concrete definition of relational inductive bias than the conceptual one proposed by \citet{battaglia2018relational}, \ie{} constraints on relationships and interactions among entities in a learning process.
Neural architectures typically encode two types of constraints: sparse connectivity and sharing parameters.
Sparse connectivity can be achieved by setting some of the $\mb{A}$ components to zero.
For example, $\mb{A}_{ab} = \mb{0}$ means no message can be passed from entity $b$ to entity $a$.
Parameter sharing simply sets some of the message passing matrices to correspond to the same matrix.
Specific patterns of connectivity and parameter sharing encode different relational inductive biases, ensuring specific, desirable properties during computation, represented by message passing operations.
For instance, the locality of a convolutional layer can be accomplished by keeping connectivity from $a$ to $b$ only if entity $a$ is in the receptive field of entity $b$.
Leveraging the~\citep[Lemma 3]{zaheer2017deep}, we can also implement the permutation equivariance (the inductive bias of Deep Set~\cite{zaheer2017deep}) by sharing parameters across diagonal submatrices and sharing parameters among the off-diagonal submatrices.

The relational graph of R-GCN provides a natural way to describe arbitrary connectivity and parameter sharing constraints.
By rearranging and applying the distributive law, we can express the message passing matrix as:
\begin{align}
    \mb{A}_{ab} = \sum_{r \in \mathcal{R}_{ba}} \frac{1}{c_{a,r}} \mb{W}_r,
    \label{eq:rel}
\end{align}
where $\mathcal{R}_{ba}$ is the set of all relation labels for relations from $b$ to $a$, and $c_{a,r},\mb{W}_r$ are the normalization constant and weight matrix of R-GCN in \cref{eq:simple}.

We now provide an example demonstrating the equivalence of R-GCN with local directional relations and 2D convolution:
The update rule for a single feature vector of a linear convolution layer can be written as
$\mb{y}_a = \sum_{b \in K_a} \mb{A}_{ab} \mb{x}_b$
, where $K_a$ is the set of entities in the receptive field around $a$ and $\mb{A}_{ab}$ is the same value for all $a$ and $b$ with the same relative position to each other.
Given \label{eq:rel}, the equivalence between R-GCN with local directional encoding and image convolution becomes clear. The only difference is that R-GCN introduces an extra normalization factor, which is a constant in this case, as most of the nodes have the same number of incoming edges.

%namely:
%\begin{align}
%    \mb{A}_{ab} = \mb{A}_{cd} \Longleftarrow (x_a-x_b=x_c-x_d)\wedge(y_a-y_b=y_c-y_d)
%\end{align}

%To express this (equivalence) constraint, each $\mathcal{R}_{ab}$ corresponds to a unique local direction relation (or the self loop), that is:
%\begin{align}
 %   \mathcal{R}_{ba} = \mathcal{R}_{dc} \iff (x_a-x_b=x_c-x_d)\wedge(y_a-y_b=y_c-y_d)
  %  \label{eq:conv_rel}
%\end{align}
%
%\zhengyao{Do we want to further expand this?}

\subsection{Relational Inductive Biases of %Image 
Convolutional Layers} \label{sec:cnn}
We now consider the case where entities are associated with a 2D feature map and analyze the relational inductive biases of convolutional layers.
For all tasks we consider, each environment observation is (or contains) a \emph{feature map}, defined as a mapping $m : W \times H \to P$ with $W = \set{0, ... ,w}$, $H=\set{0, ... ,h}$ and $P=\mathbb{R}^m$, where $w$ and $h$ is the width and height of the input, and $n$ is the dimensionality of the feature vector.
Each feature vector corresponds to an entity and therefore, a node in the relation graph.
When dealing with a feature map, a single 2D convolution layer has the following relational inductive biases:
\begin{inparaenum}[\itshape i\upshape)]
\item \emph{Locality}: Information is only transmitted from nodes in a kernel-sized receptive to the central node. Formally, $A_{ab}$ is not a zero matrix only if $a$ is in the reception field around $b$.
\item \emph{Anisotropy}: The the propagation of information in different directions follows different rules. Using our block matrix formalization: in a reception field around entity $a$, the value of $A_{ab}$ for each distinct $b$ is distinct.
\item \emph{Spatial translation equivariance}: Node pairs with the same relative position share the same message passing rules. Namely, $A_{ab}$ and $A_{cd}$ should be the same if the relative position between $a,b$ and that between $c,d$ are the same.
\end{inparaenum}
Locality is a useful relational inductive bias for most environments where feature maps abstract (or are sampled from) the physical world because the objects that are close to each other are more likely to inform or interact with each other.
For example, in the grid world tasks we considered, most interactions happen locally and even remote interactions typically rely on local, intermediary objects (\eg{} an intermediary object thrown by entity $a$ to a remote entity $b$).
Note that spatially adjacent objects may be further abstracted as a single larger entity.

Anisotropy can be used to judge directionality, useful in the RL environments considered in this work, as actions are bound to directions \eg{``\emph{move left}'' and ``\emph{move right}''}..
We emphasize this inductive bias to highlight why relations should be directional in GTG. Note that MLPs are by default anisotropic
Translation equivariance is an important inductive bias which captures the intuition that interactions between two entities likely depended on the relative positions between these two entities rather than their absolute position.
Translation equivariance relates to the assumption commonly made by feature engineering in image processing, where and the rules for extracting features should be independent of the feature position. Here, each filter corresponds to a high-level feature vector.

\subsection{Computational Costs}
The computational cost of our model is proportional to the number of ground atoms (\ie{edges in the KB and thus in the computational graph}) induced by the underlying relations. Let N be the number of entities (\eg{pixels}). Our model's time complexity ranges from $O(N)$ for convolutional layers to $O(N^2)$ for fully-connected graphs, \eg{Transformer layers}. For GTG's relation determination rules, the time complexity of local directional relations is $O(N)$, that of alignment relations (defined in \cref{sec:align}) is $O(N^{3/2})$, and that of remote relations is $O(N^2)$. 
%Additionally, as duplicated computation occurs under alignment and remote relations, more complex implementations can further reduce the time complexity.

\end{document}